%% file: main.tex
\documentclass{article} % For LaTeX2e
\usepackage[authoryear,round]{natbib}
\usepackage[preprint]{main}

\usepackage{microtype}
\usepackage{hyperref}
\usepackage{url}
\usepackage{booktabs}
\usepackage{svg}
\usepackage{amssymb} % Or
\usepackage{bbding}

\usepackage{lineno}
\usepackage{graphicx}
\usepackage{amsmath}
\usepackage{enumitem}
\usepackage{geometry}
\usepackage{xcolor}
\usepackage{xspace}
\usepackage{graphicx}
\usepackage{array} % for m{} column type (vertical centering)
\usepackage{subcaption}
\usepackage{multirow}
\usepackage{xcolor}
\usepackage{colortbl}

\newif\ifshowvd
\showvdtrue    % <- change to \showvdfalse to hide green edits

\usepackage[colorinlistoftodos, textsize=scriptsize]{todonotes}

\newcommand{\beyondweb}{%
  \textcolor[HTML]{002ECF}{\textbf{Beyond}\normalfont{Web}}\xspace
}

\definecolor{darkblue}{rgb}{0, 0, 0.5}
\definecolor{beyondwebbluedark}{HTML}{002ECF}
\definecolor{beyondwebbluelight}{HTML}{E8FBFE}
\hypersetup{colorlinks=true, citecolor=darkblue, linkcolor=darkblue, urlcolor=darkblue}
\definecolor{lightblue}{RGB}{173,216,230}
\definecolor{customlightblue}{HTML}{007cfa}
\usepackage{soul} % for the command \hl

\title{
\vspace{-1em}
\centering
\begin{tabular}{c}
\raisebox{-0.6em}{\includegraphics[height=1.1cm]{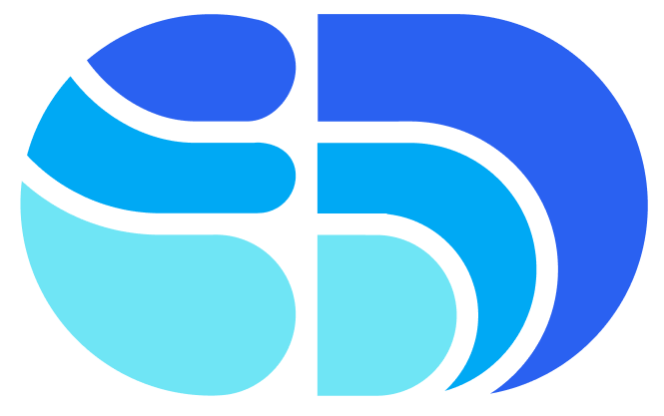}} 
\hspace{-0.05em}
\huge\beyondweb
\hspace{1em}\\[0.5em]
\Large Lessons from Scaling Synthetic Data\\ for Trillion-scale Pretraining
\end{tabular}
}
\definecolor{LightCyan}{rgb}{0.88,1,1}
\usepackage[most]{tcolorbox} % Required for colored boxes

\usepackage{tabularx}   % For tables with auto-wrapping columns
\usepackage{array}      % For custom column types
\usepackage{ragged2e}   % For better text alignment

\newcolumntype{L}{>{\RaggedRight\arraybackslash}X}

\begin{document}

\ifcolmsubmission
\linenumbers
\fi

\maketitle
\vspace{-5.75em}
\begin{center}
\textbf{DatologyAI Team}\footnotemark
\end{center}
\footnotetext{See Contributions and Acknowledgements (\S~\ref{sec:contri}) for full author list.}
% \vspace{0.75em}

\begin{abstract}
\vspace{-0.5em}
Recent advances in large language model (LLM) pretraining have shown that simply scaling data quantity eventually leads to diminishing returns, hitting a \textit{data wall}. 
In response, the use of synthetic data for pretraining has emerged as a promising paradigm for pushing the frontier of performance. 
Despite this, the factors affecting synthetic data quality remain poorly understood. In this work, we introduce \beyondweb, a synthetic data generation framework that produces high-quality synthetic data for pretraining. \beyondweb significantly extends the capabilities of traditional web-scale datasets, outperforming state-of-the-art synthetic pretraining datasets such as Cosmopedia and Nemotron-CC's high-quality synthetic subset (Nemotron-Synth) by up to \textcolor{blue}{5.1 percentage points} (pp) and \textcolor{blue}{2.6pp}, respectively, when averaged across a suite of 14 benchmark evaluations. It delivers up to \textcolor{blue}{7.7$\times$ faster} training than open web data and \textcolor{blue}{2.7$\times$ faster} than Nemotron-Synth.
Remarkably, a 3B model trained for 180B tokens on \beyondweb  outperforms an 8B model trained for the same token budget on Cosmopedia.
We also present several insights from \beyondweb\ on synthetic data for pretraining: what drives its benefits, which data to rephrase and how, and the impact of model size and family on data quality.
Overall, our work shows that there's no silver bullet for generating high-quality synthetic pretraining data. The best outcomes require jointly optimizing many factors, a challenging task that requires rigorous science and practical expertise. Naive approaches can yield modest improvements, potentially at great cost, while well-executed methods can yield transformative improvements, as exemplified by \beyondweb.
\end{abstract}

\vspace{-1em}
\begin{figure}[!h]
    \centering
    \includegraphics[width=0.88\textwidth]{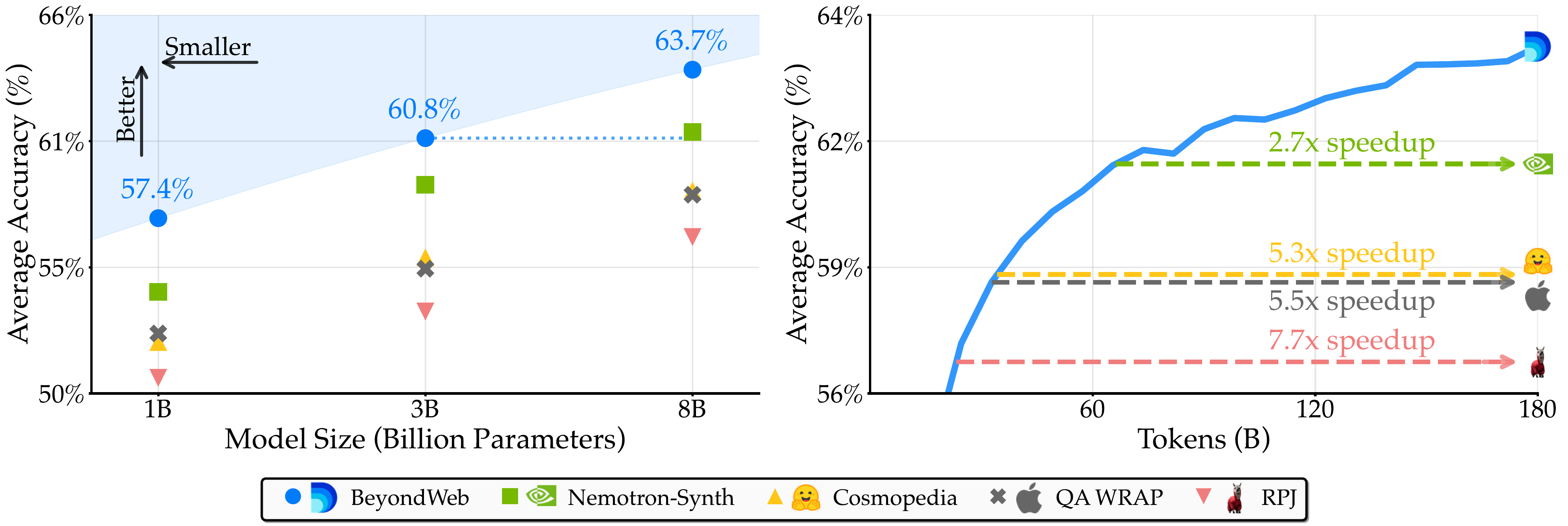}
    \caption{
    \textbf{Left:} \beyondweb establishes a new Pareto frontier for synthetic pretraining data. 
    Notably, our 3B model outperforms all but one 8B model trained on baseline datasets with the same token budget.
    Average Accuracy (\%) is the mean across 14 benchmarks. 1B model trained for 1T tokens; 3B and 8B models for 180B tokens. 
    \textbf{Right:} For 8B model, we achieve up to 7.7$\times$ and 2.7$\times$ speedup (in time to reach baseline accuracy) over RedPajama and Nemotron-Synth respectively.}
    \label{fig:performance_overview}
\end{figure}

\input{sections/1-intro}
\input{sections/2-background}
\input{sections/3-beyondweb}

\input{sections/4-synth_data_eval}
\input{sections/5-future}
\input{sections/Conclusion}

\bibliographystyle{abbrvnat}  % Abbreviated author names with year
\bibliography{main}

\appendix
\input{sections/appendix}

\end{document}

%% file: sections/1-intro.tex
\section{Introduction}
\label{sec:intro}

Until 2024, breakthroughs in language modeling followed a predictable recipe: train ever-larger models on exponentially larger internet-scraped datasets. However, as the scale of data collection ballooned into the trillions of tokens, the field began to encounter a \textit{data wall}, beyond which data of high information-density becomes prohibitively scarce. The returns on collecting more internet data rapidly diminished, pushing researchers toward exploring alternative paradigms. 

Synthetic data has emerged as a powerful complement to scarce, high-quality web text. Initial evidence from \textbf{Tiny Stories}~\citep{eldan2023tinystories} showed that targeted prompting of large language models can generate data suitable to train small language models from scratch. Follow-up work, most notably Microsoft’s \textbf{Phi} family~\citep{li2023textbooks} and the open-source \textbf{Cosmopedia} corpus~\citep{benallal2024cosmopedia}, demonstrated that sub-2B models jointly trained on synthetic and raw data can outperform much larger baselines. We refer to this practice of using large models to generate training data \textit{de novo} as the \textbf{generator-driven} approach.

Generator-driven approaches, while powerful, are ultimately limited by the cost and idiosyncrasies of the large generator models they rely on, such as GPT-4~\citep{maini_phi_1_5}. To address these limitations, the Web Rephrase Augmented Pre-training \textbf{(WRAP)} paradigm~\citep{maini2024rephrasing} and its refinement in \textbf{Nemotron-CC}~\citep{su2024nemotron} developed what we refer to as the \textbf{source rephrasing} approach. In these works, rather than prompting a large LLM to create knowledge de novo, small LLMs are used to rephrase existing web data into higher-quality, task-aligned formats (e.g., Q\&A pairs, instructional passages). This approach achieves superior coverage and diversity at lower compute costs, establishing synthetic rephrasing as a practical solution for pretraining-scale data generation.

The success of synthetic data methods has led research organizations to pour substantial compute resources into creating ever larger corpora of rephrased synthetic data~\citep{kimiteam2025kimik2openagentic,hui2024qwen2,musk2024tweet}. Most recently, synthetic data was specifically highlighted as a key innovation in GPT-5's development~\citep{gpt5-livestream}. While we know that synthetic pretraining data \emph{can} work, we still lack a comprehensive scientific understanding of the factors that determine \emph{when} and \emph{how} synthetic pretraining data works.
Our work systematically addresses these questions and highlights the challenges of scalably generating high-quality pretraining data by conducting experiments across model scales up to 8B parameters. Specifically, we address the following:

\textbf{How Does Synthetic Data Provide its Benefits?}
\vspace{-.7em}
\begin{itemize}[leftmargin=*, labelsep=0.2em]
     \item First, we assess whether the model quality improvements imparted by the \emph{generator-driven} paradigm can be explained as distillation of a teacher model's knowledge into higher per-token information density data. We find that simple summarization prompts in the \emph{source-rephrasing} paradigm can match the performance of generator-driven approaches such as Cosmopedia (\S\ref{sec:knowledge_transfer}).
    \item Next, we ask whether synthetic data can actually breach the \textit{data wall}. We show that in a data-constrained setting, naive approaches such as simple continuations (of existing web data) provide limited accuracy improvements over just repeating data. However, thoughtfully-created data that \emph{fills in} the distributional gaps of the web data has much larger benefits (\S\ref{sec:data_wall}).
\end{itemize}

\textbf{What and How to Rephrase?}
\vspace{-.7em}
\begin{itemize}[leftmargin=*, labelsep=0.2em]
    \item First, data \textbf{quality} matters: rephrasing high-quality data provides benefits over using lower-quality sources, but high-quality input data alone is not enough to yield the highest quality synthetic data (\S\ref{sec:quality_matters}).
    \item Second, distributional \textbf{style} matching is important. Web data contains only 2.7\% conversational content, despite chat being a major inference use case. Style matching improves performance in downstream tasks, but the benefits saturate quickly with the proportion of conversational data, indicating that it is not a sufficient solution. (\S\ref{sec:style_matters}).
    \item Third, \textbf{diversity} in generation strategies is critical to leverage continued benefits from synthetic data when scaling to large training budgets of trillions of tokens (\S\ref{sec:diversity}).
\end{itemize}

\textbf{Who Should Rephrase?}
\vspace{-.7em}
\begin{itemize}[leftmargin=*, labelsep=0.2em]
    \item Synthetic data benefits are largely consistent across different rephraser families, and rephraser quality does not predict synthetic data quality, indicating that rephrasing is a generic capability and rephraser model efficacy is difficult to predict.(\S\ref{sec:family_consistency}).
    \item We observe diminishing returns when increasing rephraser size beyond 3B parameters, with 8B LLMs achieving only marginal gains over 3B. This validates that (given the right recipe) effective synthetic data generation doesn't necessarily require massive computational resources (\S\ref{sec:generator_scaling}).
\end{itemize}

Motivated by these insights, we introduce \beyondweb, a synthetic pretraining data generation paradigm that leverages targeted document rephrasing to yield diverse, relevant, and information-dense synthetic data. On 8B models trained for 180B tokens, we outperform RedPajama \citep{weber2024redpajama} and the high-quality synthetic subset of Nemotron-CC (Nemotron-Synth) by \textcolor{blue}{+7.1pp} and \textcolor{blue}{+2.6pp}, respectively, while achieving training speedups of \textcolor{blue}{7.7x} and \textcolor{blue}{2.7x} (see Figure~\ref{fig:performance_overview}). Futhermore, 3B model trained on \beyondweb surpasses the performance of strong 8B baselines such as Cosmopedia. This establishes a new \textbf{pareto frontier} for the accuracy-efficiency trade-off in LLM pretraining.
Our results also indicate a promising scaling trend: the benefits of our approach show consistent gains across model sizes, achieving \textcolor{blue}{(+3.1pp, +2.0pp, +2.6pp)} over Nemotron-Synth at 1B, 3B, 8B parameters respectively.

We note that \beyondweb is one part of our full curation platform, which was used to curate the 7T token pretraining dataset for ArceeAI's AFM4.5B model \citep{atkins2025afm45b}—and combining \beyondweb with our full curation platform obtains even better results. The strength of AFM4.5B demonstrates the efficacy of \beyondweb not just in the controlled experimental settings presented here, but also as part of an end-to-end curation pipeline for generating foundation-scale pretraining datasets for production models.

Overall, our work demonstrates that generating high quality synthetic data is not a trivial task, and that there is no singular silver bullet solution for obtaining the highest quality synthetic data. Naive approaches to synthetic data generation can provide little benefit or even some detriment, while also incurring substantial computational cost for generation. In contrast, the right synthetic generation approach can yield exceptional model quality improvements, as demonstrated by our \beyondweb results. Getting synthetic data right requires careful consideration of numerous factors, including data selection strategies, generation methodologies, diversity preservation, and quality control, making it a challenging endeavor that demands rigorous science and practical expertise.

%% file: sections/2-background.tex
\section{A Tale of Two Approaches for Synthetic Pretraining Data}

\subsection{The Generator-Driven Paradigm: Creating Knowledge from Models \emph{de Novo}}

The \textbf{generator-driven} approach to synthetic data uses large-scale models to generate training data from scratch, encapsulating the knowledge embedded within these models. By generating data from a pretrained model, researchers have been able to substantially improve the performance of the next generation of models. Notably, the seminal work on \textbf{Tiny Stories}~\citep{eldan2023tinystories} demonstrated that high-quality, carefully prompted synthetic data, such as simplified narratives crafted by GPT-4, could effectively train small language models from scratch. This result defied the prevailing trend of scaling models and web data, sparking a paradigm shift and paving the way for increased use of synthetic data for pretraining small, performant models.

Soon after, the Phi model family~\citep{li2023textbooks} advanced the generator-driven approach by training small ($<$2B parameter) models jointly on synthetic data and raw web data, and were able to outperform models up to $10\times$ larger using a fraction of the training compute. \textbf{Cosmopedia}~\citep{benallal2024cosmopedia} furthered this direction by introducing a large-scale, open-source synthetic dataset, generated by leveraging open-source LLMs and prompting it with a diverse set of curated seed topics. However, this paradigm, which relies on using a powerful existing model as a knowledge bank to generate synthetic data, is constrained by the high computational cost of accessing state-of-the-art generators like GPT-4, making it inaccessible to many researchers. Furthermore, these approaches do not scale well and are susceptible to model collapse \citep{shumailov2024curserecursiontraininggenerated, briesch2024largelanguagemodelssuffer, guo2024curiousdeclinelinguisticdiversity}, as they inherit the knowledge and biases of the generator, potentially resulting in limited diversity, coverage gaps, and hallucinated content~\citep{maini_phi_1_5}.

\subsection{The Source Rephrasing Paradigm: Enhancing Existing Knowledge}

Motivated by the shortcomings of the generator-driven approach, \citet{maini2024rephrasing} recently proposed an alternative synthetic generation methodology: \textbf{Web Rephrase Augmented Pre-training (WRAP)}. WRAP leverages existing web documents and employs smaller models to rephrase this content into higher-quality, structured and/or targeted formats. Conditioning on existing data significantly reduces dependence on expensive large models while enriching the training corpus with natural, diverse, and topical language. \citet{maini2024rephrasing} were able to speed up pretraining by more than 3$\times$ through the use of strategic rephrasing.
Of note, \citet{allen2023physics} concurrently worked on rephrasing articles during pretraining on a toy task of author biographies, and showed such rewriting enhanced model performance. Recently, \textbf{REWIRE}~\citep{nguyen2025recyclingwebmethodenhance} proposed rewriting medium-quality documents into higher-quality ones, which would otherwise be filtered out by classifiers, thereby augmenting the high-quality pool.
We refer to this collection of methodologies as the \textbf{source rephrasing} paradigm.

The source rephrasing approach has gained considerable traction in industry. Projects such as Nvidia's Nemotron-CC~\citep{su2024nemotron}, StabilityAI's multilingual augmentation efforts~\citep{pieler2024rephrasing}, and Microsoft's Phi-4~\citep{abdin2024phi} have all built on this paradigm, emphasizing the value of augmenting real-world internet data. This highlights a critical ideological shift toward data-driven rather than model-driven synthetic data generation. This approach effectively combines the advantages of broad, naturally occurring knowledge from the internet with controlled style and format enhancements. In 2025, rephrasing has become the dominant paradigm with state-of-the-art LLMs like Kimi K2~\citep{kimiteam2025kimik2openagentic}, Qwen-2.5~\citep{hui2024qwen2}, Grok~\citep{musk2024tweet}, and GPT-5 \citep{gpt5-livestream} reporting substantial use of and/or meaningful gains from source rephrasing.

\subsection{Synthetic Data Beyond Pretraining}

The utility of synthetic data extends beyond pretraining and into other phases of model development, including fine-tuning and evaluation. Synthetic data can play a pivotal role in addressing specific task-oriented training objectives, aligning model behavior, reducing toxicity, and improving generalization on targeted downstream tasks. Techniques such as instruction-tuning, Chain-of-Thought prompting, and data backtracing emphasize synthetic data's potential beyond pretraining by directly influencing downstream capabilities, such as reasoning \citep{lu-etal-2024-mathgenie}, alignment \citep{li2024selfalignmentinstructionbacktranslation, wang-etal-2024-codeclm}, and reducing model hallucinations \citep{jones2024teaching}. However, the relative lack of synthetic data research for pretraining compared to post-training and the potential magnitude of impact from successful pretraining improvements motivated us to focus exclusively on synthetic data for pretraining in this work.

%% file: sections/3-beyondweb.tex
\section{Introducing \beyondweb}
\label{sec:beyondweb}

We introduce \beyondweb, a synthetic data approach that leverages grounding and diversity to substantially improve language model pretraining efficiency. It combines the broad coverage of web-scale corpora with strategically-generated content that fills critical gaps, particularly in underrepresented styles, formats, and topics. \beyondweb employs diverse generation strategies, including format transformation (e.g. converting web content into question–answer pairs), style modification (e.g. enhancing pedagogical tone), and content restructuring to improve information density and accessibility. This enables the creation of models better aligned with real-world language use and downstream task demands.

\paragraph{Datasets.} We compare \beyondweb against state-of-the-art public synthetic pretraining datasets and methodologies, as well as a non-synthetic open web dataset, RedPajama (RPJ)~\citep{weber2024redpajama}. We chose RPJ because it is a well-established baseline that has minimal curation applied to it, allowing us to robustly assess the effects synthetic data. 
For comparisons with other synthetic datasets, we adopt a mixed curation strategy: 60\% of the training tokens are sourced randomly from RPJ, while 40\% come from synthetic data. We evaluate against three synthetic baselines: 
\begin{itemize}[leftmargin=*, labelsep=0.2em]
    \item \textbf{Cosmopedia}~\citep{benallal2024cosmopedia} contains over 39 million textbooks, blog posts, and stories generated by Mixtral-8x7B-Instruct-v0.1, using web-derived prompts covering a predefined set of topics. We use the latest version, Cosmopedia-v2, which contains approximately 27 billion tokens. When more tokens are needed for our experiments, we repeat the dataset to compensate. 
    While repetition is not ideal because it potentially reduces the effective per-token quality of the dataset, we consider this a general limitation of the generator-driven paradigm that stems from the cost and difficulty of prompting and generation. 
    \item \textbf{WRAP}~\citep{maini2024rephrasing} rephrases web content into various formats, with question-answering being the most performant style. For our comparison, we use the RedPajama dataset as the source corpus for rephrasing, and the Llama-3.1-8b-Instruct LLM for rephrasing.
    \item \textbf{Nemotron-Synth}~\citep{su2024nemotron} is the high-quality subset of synthetic data in Nemotron-CC. The data was generated by applying diverse rephrasing prompts to high-quality, classifier-selected inputs. We chose the subset of Nemotron-CC synthetic data derived from high-quality input data, which we refer to as Nemotron-Synth, instead of the full Nemotron-CC synthetic data because Nemotron-Synth is an extremely competitive baseline and sufficiently large for our long-horizon experiments. Nemotron-Synth contains 1.5 trillion tokens, so we randomly sample a subset while maintaining the original proportions across different data subsets. 
\end{itemize}

We apply our \beyondweb approach to a high-quality subset of DCLM~\citep{li2024datacomplm} selected using the methods described in \citet{datologyai_text_2024}. 

\paragraph{Synthetic Generation Infrastructure.}
Generating foundation-scale synthetic pretraining data presents a formidable engineering challenge. We addressed this by building a massively scalable data curation pipeline that collects billions of high-quality documents suitable for rephrasing (as detailed in \citealt{datologyai_text_2024}) and then efficiently performs large-scale parallelizable synthetic rephrasing over the resulting trillions of tokens. Furthermore, because our curation platform is not only central to our internal research but also serves as our product, it was designed to be deployable across heterogeneous customer environments. Our initial setup used Slurm on an AWS Hyperpod cluster of H100s, but this system was rigid, slow to iterate on, had deployment constraints, and made it difficult to track experiment choices. This motivated us to move to a solution using Ray and vLLM on Kubernetes. The new setup supports heterogeneous clusters and various GPU types, scales efficiently on Kubernetes, and integrates seamlessly into our existing Ray and Spark curation pipeline. In addition to facilitating deployment and reducing cost and complexity, this also enables us better end-to-end experiment tracking. This infrastructure shift has transformed our synthetic generation system into a flexible, scalable, and production-ready platform that simultaneously accelerates research, streamlines deployment, and powers customer-facing workloads. We look forward to detailing our synthetic data generation infrastructure in a follow-up release.

\paragraph{Training setup.}
We train three sizes of LLMs: a 1B parameter model trained on 1 trillion tokens, a 3B model trained on 180 billion tokens, and an 8B model trained on 180 billion tokens. We use \textsc{LLaMA-3.2} architecture \citep{grattafiori2024llama} for the 1B and 3B models, and \textsc{LLaMA-3.1} architecture for the 8B model. Additional training details can be found in Appendix \ref{sec:app_training}. We did not extensively tune hyperparameters other than an early search on the baseline RPJ data, as our goal was to improve model performance with data interventions alone.

\paragraph{Evaluation setup.} We evaluated the models on 14 benchmark tasks (enumerated in Appendix \ref{sec:eval_tasks}) using both 0-shot and 5-shot prompting and report average accuracy across all settings and tasks. Multiple-choice questions are assessed using a relative scoring method, as described in Hugging Face's analysis of the OpenLLM leaderboard and referred to as "cloze form" (CF) in \citet{gu2025olmes}. This approach compares the probabilities assigned by the model, restricted to the set of valid answer choices.

\paragraph{Performance improvements across scales.}
\beyondweb demonstrates consistent improvements across all evaluated model sizes, as shown in Figure~\ref{fig:performance_overview}. At 1B parameters, we achieve 57.4\% average accuracy (\textcolor{blue}{\small+6.7pp} over RPJ), establishing a strong foundation. The performance gains become more pronounced as scale increases: at 3B parameters, we reach 60.8\% accuracy (\textcolor{blue}{\small+7.3pp} over RPJ), and at 8B parameters, accuracy rises to 63.7\% (\textcolor{blue}{\small+7.1pp} over RPJ). \beyondweb also consistently outperforms Nemotron-Synth, our strongest synthetic pretraining data baseline, at all scales (\textcolor{blue}{+3.1pp, +2.0pp, +2.6pp} at 1B, 3B, and 8B, respectively). These results demonstrate that our synthetic data approach remains effective across scales and continues to offer substantial benefits in larger models.

\paragraph{Train Models 7.7x faster.}

\beyondweb enables substantial computational savings during pretraining, as shown in Figure~\ref{fig:performance_overview}. For the 8B model, \beyondweb matches or exceeds RedPajama’s 180B-token performance in just 23.2B tokens (7.7× speedup) and Nemotron-Synth’s 180B-token performance in only 66.2B tokens (2.7× speedup). Faster convergence directly reduces training cost, energy consumption, and iteration time, affording more experiments within a fixed compute budget. This efficiency is valuable not only for large industry labs aiming to optimize costs but also for relatively smaller organizations constrained by infrastructure, thereby helping to democratize access to high-performance LLMs.

\paragraph{Establishing a new pareto frontier for synthetic data.}
Our results establish a new Pareto frontier for the speed-accuracy trade-off in language model pretraining, as visualized in Figure~\ref{fig:performance_overview}. Remarkably, our 3B model (60.8\% accuracy) outperforms all but the strongest of the 8B baseline models trained for the same number of tokens (180B), demonstrating that high-quality synthetic data can achieve superior results with dramatically fewer parameters. This highlights the scalability and headroom unlocked by high-quality synthetic data, enabling stronger models at lower computational cost and challenging the conventional reliance on ever-larger architectures.

\input{tables/hero_avgshot}

\paragraph{Consistent improvements across tasks.} As presented in Table~\ref{tab:tasks_hero_avgshot}, \beyondweb achieves the highest average accuracy at every model scale and also outperforms baselines on 13, 12, and 13 out of 14 tasks, in 1B, 3B, and 8B models, respectively. This demonstrates that the performance gains are not confined to a few standout benchmarks but are broadly distributed, reflecting the generalizability enabled by high-quality, diverse synthetic data.

We now examine the factors that contribute to the strength of synthetic data approaches such as \beyondweb and the obstacles to generating high-quality synthetic pretraining data.

%% file: tables/hero_avgshot.tex
% Single table with Scale | Method | Task1 | Task2 | ... | Task15
\begin{table}[htbp]
\centering
\resizebox{\textwidth}{!}{
\begin{tabular}{ll|ccccccccccccccc}
\toprule
\textbf{Scale} & \textbf{Dataset} & \textbf{ARC(C)} & \textbf{ARC(E)} & \textbf{BoolQ} & \textbf{COPA} & \textbf{CSQA} & \textbf{Hella.} & \textbf{MMLU} & \textbf{OBQA} & \textbf{PIQA} & \textbf{RACE-H} & \textbf{RACE-M} & \textbf{SIQA} & \textbf{SciQ} & \textbf{Wino.} & \textbf{Avg.} \\
\toprule
\multirow{5}{*}{\parbox{1cm}{\centering \textbf{1B} \\ \textbf{(1TT)}}} & RPJ & 29.4 & 55.9 & 60.0 & 72.0 & 43.8 & 51.8 & 31.5 & 35.2 & 71.3 & 35.4 & 43.1 & 43.2 & 82.8 & 53.9 & 50.7 \\
 & QA WRAP & 30.9 & 59.6 & \underline{65.7} & \underline{74.5} & 44.2 & 51.1 & 32.7 & 35.2 & 71.8 & 37.0 & 47.0 & 44.3 & \underline{88.2} & 53.3 & 52.5 \\
 & Cosmopedia & 34.0 & 62.8 & 59.7 & 70.5 & 43.9 & 55.6 & 32.8 & \underline{37.0} & 73.3 & 34.5 & 43.4 & \underline{45.4} & 84.4 & 53.1 & 52.2 \\
 & Nemotron-Synth & \underline{35.9} & \underline{64.2} & 62.8 & 71.5 & \underline{47.1} & \textbf{56.2} & \underline{34.6} & 36.5 & \underline{74.1} & \underline{39.3} & \underline{52.6} & 44.3 & 86.9 & \underline{54.1} & \underline{54.3} \\
 & BeyondWeb & \colorbox{lightblue}{\textbf{41.4}} & \colorbox{lightblue}{\textbf{70.7}} & \colorbox{lightblue}{\textbf{68.3}} & \colorbox{lightblue}{\textbf{76.5}} & \colorbox{lightblue}{\textbf{49.3}} & \underline{56.1} & \colorbox{lightblue}{\textbf{35.5}} & \colorbox{lightblue}{\textbf{39.1}} & \colorbox{lightblue}{\textbf{75.0}} & \colorbox{lightblue}{\textbf{41.7}} & \colorbox{lightblue}{\textbf{53.7}} & \colorbox{lightblue}{\textbf{47.4}} & \colorbox{lightblue}{\textbf{93.0}} & \colorbox{lightblue}{\textbf{56.0}} & \colorbox{lightblue}{\textbf{57.4}} \\
\midrule
\multirow{5}{*}{\parbox{1cm}{\centering \textbf{3B} \\ \textbf{(180BT)}}} & RPJ & 33.0 & 61.3 & 61.5 & 70.0 & 47.9 & 58.6 & 33.8 & 36.1 & 74.6 & 37.7 & 46.7 & 44.7 & 87.4 & 55.1 & 53.5 \\
 & QA WRAP & 36.0 & 64.4 & \underline{67.2} & 73.0 & 46.5 & 58.6 & 35.0 & 36.5 & 73.2 & 40.1 & 51.7 & 46.2 & 90.1 & 55.2 & 55.3 \\
 & Cosmopedia & 37.5 & 66.1 & 63.1 & 74.5 & 48.9 & 62.3 & 35.5 & 39.1 & 76.0 & 38.4 & 49.3 & 46.2 & 88.0 & 55.5 & 55.8 \\
 & Nemotron-Synth & \underline{41.2} & \underline{71.4} & 66.3 & \underline{80.5} & \underline{50.2} & \textbf{63.9} & \underline{37.7} & \underline{41.4} & \underline{76.8} & \underline{42.7} & \underline{55.0} & \underline{46.5} & \underline{92.2} & \textbf{57.7} & \underline{58.8} \\
 & BeyondWeb & \colorbox{lightblue}{\textbf{46.7}} & \colorbox{lightblue}{\textbf{75.3}} & \colorbox{lightblue}{\textbf{70.8}} & \colorbox{lightblue}{\textbf{82.0}} & \colorbox{lightblue}{\textbf{51.7}} & \underline{63.5} & \colorbox{lightblue}{\textbf{37.7}} & \colorbox{lightblue}{\textbf{43.0}} & \colorbox{lightblue}{\textbf{77.0}} & \colorbox{lightblue}{\textbf{44.6}} & \colorbox{lightblue}{\textbf{57.9}} & \colorbox{lightblue}{\textbf{49.4}} & \colorbox{lightblue}{\textbf{94.0}} & \underline{57.5} & \colorbox{lightblue}{\textbf{60.8}} \\
\midrule
\multirow{5}{*}{\parbox{1cm}{\centering \textbf{8B} \\ \textbf{(180BT)}}} & RPJ & 35.0 & 65.9 & 66.1 & 75.5 & 51.4 & 65.7 & 36.0 & 39.5 & 76.3 & 39.6 & 47.7 & 47.0 & 89.2 & 57.7 & 56.6 \\
 & QA WRAP & 39.9 & 68.4 & \underline{70.5} & 79.0 & 50.4 & 64.3 & 37.8 & 38.3 & 75.5 & 41.5 & 54.5 & 47.0 & \underline{92.7} & 57.7 & 58.4 \\
 & Cosmopedia & 41.5 & 70.7 & 64.3 & 79.5 & 50.6 & \underline{68.2} & 37.8 & 41.6 & 77.4 & 41.2 & 52.1 & 47.3 & 89.5 & 58.4 & 58.6 \\
 & Nemotron-Synth & \underline{46.1} & \underline{75.8} & 67.9 & \textbf{83.5} & \underline{52.4} & \textbf{69.1} & \underline{40.0} & \underline{41.8} & \underline{77.6} & \underline{44.1} & \underline{57.5} & \underline{47.4} & 92.5 & \underline{58.9} & \underline{61.1} \\
 & BeyondWeb & \colorbox{lightblue}{\textbf{51.8}} & \colorbox{lightblue}{\textbf{79.3}} & \colorbox{lightblue}{\textbf{75.5}} & \underline{83.0} & \colorbox{lightblue}{\textbf{54.3}} & \colorbox{lightblue}{\textbf{69.1}} & \colorbox{lightblue}{\textbf{40.6}} & \colorbox{lightblue}{\textbf{45.6}} & \colorbox{lightblue}{\textbf{78.6}} & \colorbox{lightblue}{\textbf{47.0}} & \colorbox{lightblue}{\textbf{60.5}} & \colorbox{lightblue}{\textbf{50.3}} & \colorbox{lightblue}{\textbf{95.6}} & \colorbox{lightblue}{\textbf{60.5}} & \colorbox{lightblue}{\textbf{63.7}} \\
\bottomrule
\end{tabular}
}
\caption{Performance across all tasks averaged over 0-shot and 5-shot for different model scales and data curations. The best value at each scale is highlighted in \textbf{bold}, the second-best is \underline{underlined}, and results where \beyondweb outperforms all other methods are additionally shaded in \colorbox{lightblue}{blue}.
}
\label{tab:tasks_hero_avgshot}
\end{table}

%% file: sections/4-synth_data_eval.tex
\section{Systematically Evaluating Synthetic Data}
\label{sec:synthetic_data}

Despite the ready availability of web data for pretraining, training models solely on internet-sourced text presents several challenges. First, \textbf{data redundancy} is a major issue. Web data is finite, and modern pretraining budgets are beginning to exceed the scale of high-quality web data. Unfortunately, repeating data leads to diminishing returns and overfitting~\citep{muennighoff2023scaling,goyal2024scaling}, and thus is not a substitute for new data. The second challenge is \textbf{style mismatch}, which describes the stylistic differences between the data models are trained on and the data they encounter at test time. The top three most common styles of web data are personal blogs, product pages, and news articles~\citep{wettig2025organize}, whereas deployed models predominantly interact in conversational or instructional formats~\citep{maini2024rephrasing}. The third challenge is that \textbf{knowledge bottlenecks} emerge in underrepresented domains, making models blind to certain topics. Finally, \textbf{training efficiency} can be significantly improved if data is structured to maximize per-token knowledge utility rather than merely increasing token count.

\begin{table}[!ht]
\centering
\begin{tabular}{p{12cm} c}
\toprule
\textbf{Finding} & \textbf{Section} \\
\midrule
\multicolumn{2}{c}{\textbf{What Does Synthetic Data Really Do?}} \\
\midrule
\textbf{Are Generator-Driven Approaches Approximated by Summarization?} Simple summarization approaches that increase per-token information density achieve similar performance as Cosmopedia. But carefully-crafted source rephrasing approaches can substantially outperform both. & \S\ref{sec:knowledge_transfer} \\
\textbf{Surpassing the Data Wall}: Synthetic data generation must be thoughtful in order to breach the data wall. Naive strategies can not break the data wall.& \S\ref{sec:data_wall} \\
\midrule
\multicolumn{2}{c}{\textbf{How to Rephrase: Methods \& Techniques}} \\
\midrule
\textbf{Quality Matters}: Synthetically rephrasing high-quality web data offers larger performance gains than using low-quality web data. Nevertheless, high-quality input data alone is insufficient for generating the highest quality synthetic data. & \S\ref{sec:quality_matters} \\
\textbf{Style Matters}: Upsampling natural conversational data improves downstream task performance, though these improvements are modest and show diminishing returns with the proportion of conversational data.  & \S\ref{sec:style_matters} \\
\textbf{Diversity Matters}: As we scale synthetic data to 1T tokens, the importance of diversity in synthetic data generation is critical to avoid diminishing returns. & \S\ref{sec:diversity} \\
\midrule
\multicolumn{2}{c}{\textbf{Who Should Rephrase: Generator Properties}} \\
\midrule
\textbf{Model Family Robustness}: Synthetic data benefits are largely consistent across generator model families, and generator model quality is not predictive of synthetic data quality. Selecting a good rephrasing model may be straightforward as most work well, but selecting the best rephrasing model is not. & \S\ref{sec:family_consistency} \\
\textbf{Generator Size Saturation}: While larger generators help, improvements plateau beyond 3B parameters with diminishing returns beyond this scale. & \S\ref{sec:generator_scaling} \\
\bottomrule
\end{tabular}
\caption{Summary of key findings organized by theme. Each finding represents a core insight from our systematic evaluation of synthetic data generation strategies.}
\label{tab:key_findings}
\end{table}

These challenges highlight the need to move beyond web data by generating synthetic data. However, not all synthetically generated data are equally effective. Through extensive experimentation and systematic ablations, we examine how synthetic data can help overcome data limitations, better align training distributions with downstream use-cases, and enhance both knowledge efficiency and diversity.
Our systematic evaluation reveals several key findings, structured around three central themes: the capabilities of synthetic data, effective techniques for rephrasing, and the properties of generator models. Table~\ref{tab:key_findings} summarizes these insights along with section references.

\subsection{Experiment Setup}
\label{sec:synth_exp_setting}
As a baseline for the following sections, we use a subset of the RedPajama dataset~\citep{weber2024redpajama}. 
To construct high-quality splits from RedPajama, we follow the method from \citet{datologyai_text_2024} \textbf{(HQ Web)}.
Unless otherwise specified, all ablations train a Llama-3.2-1B model on a total budget of 20 billion tokens, with a 50:50 split between real (HQ Web) and synthetic data. 
Unless the experimental ablation demands otherwise, all synthetic data generation approaches rephrase the \textbf{same} 10 billion tokens of HQ Web data that are already a part of the final 20 billion data mixture. This means that the total \emph{source knowledge} from the internet is fixed at 10 billion tokens. For a fair baseline that controls for the amount of new knowledge, \textbf{RPJ-HQ} baseline also sees the same 10 billion tokens, but twice during its training.
For other curated datasets, such as Cosmopedia~\citep{benallal2024cosmopedia}, we randomly sample 10 billion tokens from their corpus. All evaluations report the average performance across 0-shot and 5-shot settings on a benchmark suite of 14 tasks enumerated in Appendix \ref{sec:eval_tasks}.

\subsection{Research Question (RQ) 1: Are Generator-Driven Approaches Approximated by Summarization?}
\label{sec:knowledge_transfer}

One hypothesis that may explain the benefits of synthetic data is that it effectively increases the \textbf{per-token information density} of training data. Rather than merely generating more tokens, the efficacy derives from restructuring and distilling existing knowledge into more compact and informative representations. 
This hypothesis particularly prevails in the generator-driven paradigm, in which a large LLM is seeded with various topics of interest, and asked to create textbook-style material based on its parametric knowledge. Intuitively, when a model writes a book on a specific topic, say \emph{laws of motion}, it distills all the information it has read about it on the internet into a much more condensed representation.
To test this hypothesis, we 
compare a generator-driven approach (Cosmopedia), with a simple rephrasing approach 
that merely attempts to increase the information density of the internet by \emph{summarizing} it into more compact, high-quality content.

\begin{tcolorbox}[title=Summarization Prompt]
\label{fig:summary_prompt}
Summarize the following text. Directly start with the summary. Do not say anything else.
\end{tcolorbox}

\paragraph{Experiment Design.} We consider two approaches: (1)~\textbf{Cosmopedia}, which prompts a large model to explain or teach concepts, reorganizing knowledge into compact, pedagogically optimized outputs. Cosmopedia uses an 8x7B model to generate high-quality data conditioned on seed topics. (2)~\textbf{Summarization}, which compresses existing text using a prompt that aims to preserve essential information while reducing length. This approach leverages an 8B-parameter model with a simple prompt to summarize source content. Both approaches aim to condense learning-relevant content into fewer tokens, enabling more efficient training under fixed budgets. Following the setup in Section~\ref{sec:data_wall}, we allocate 10 billion tokens for synthetic data and 10 billion tokens for the original dataset during training.

\begin{figure}[!ht]
    \centering
    \includegraphics[width=0.6\textwidth]{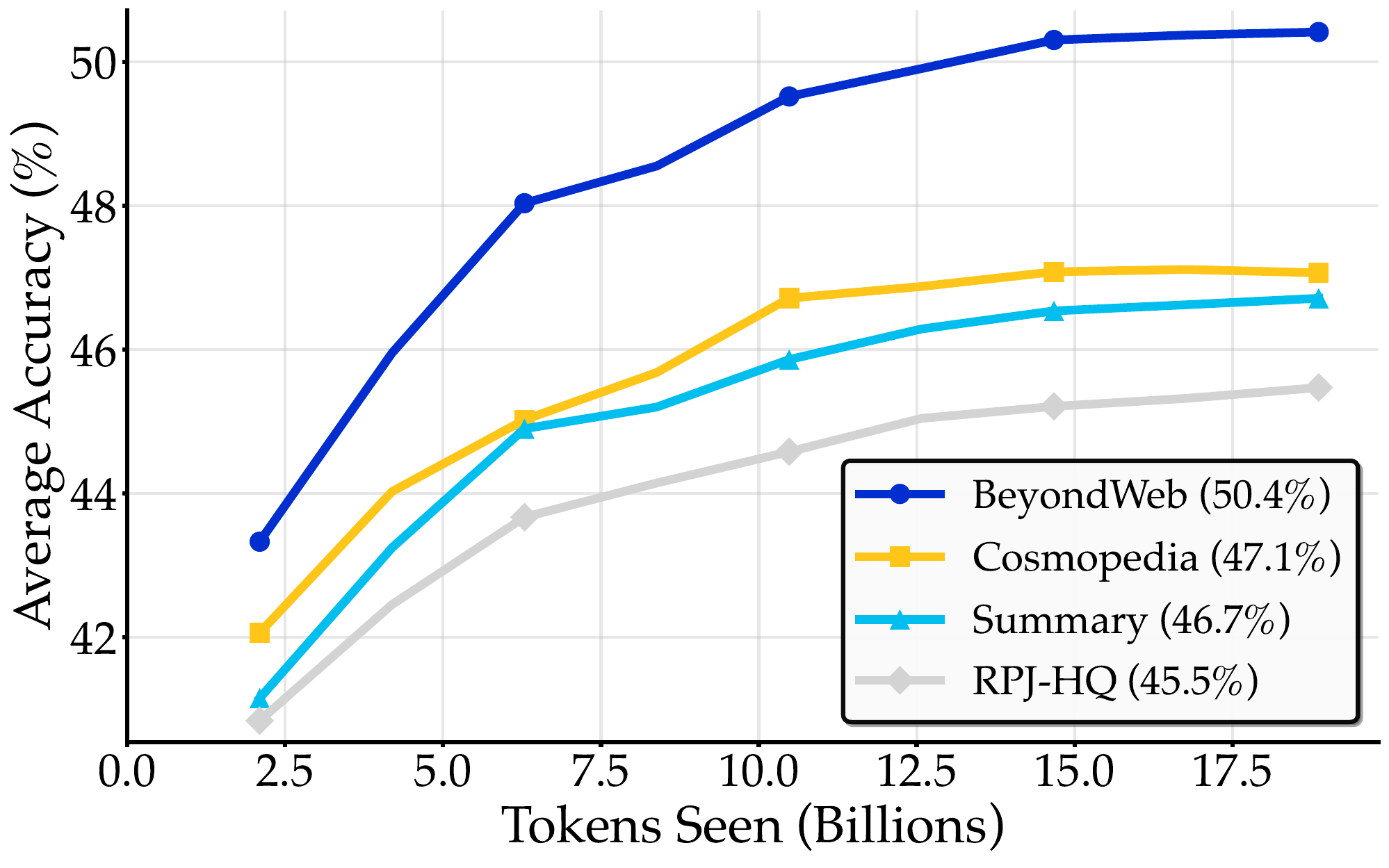}
    \caption{Knowledge transfer effectiveness across different synthetic data approaches. The yellow line represents \textbf{Cosmopedia (46.8\%)}, which uses a generator-driven, sophisticated educational content generation technique and 8x7B model; the cyan line denotes the \textbf{Summary (46.8\%)} approach, which uses an 8B model and a simple summarization prompt; the gray line denotes the \textbf{RPJ-HQ} (no synthetic data) baseline. These results demonstrate that even naive summarization achieves substantial improvements similar to those of Cosmopedia, suggesting distillation works through increased information density rather than complex knowledge transfer.}
    \label{fig:distillation_comparison}
\end{figure}

\paragraph{Results and Observations.}

Our analysis in Figure~\ref{fig:distillation_comparison} reveals that even simple distillation strategies can achieve substantial performance improvements.

\begin{enumerate}[leftmargin=*]
\item \textbf{Simple summaries match generator-driven methods}: The summarization approach (46.7\%) nearly matches the performance of the more sophisticated and compute-intensive Cosmopedia approach (47.1\%). This suggests that much of the benefit of synthetic data arises from basic information condensation. Notably, summarization, which uses a single 8B-parameter model along with information on the internet, nearly matches the performance of Cosmopedia, which relies on a much larger setup of 8$\times$7B-parameter model and manually curated seed topics to guide generation. Importantly, this shows that in the case of Cosmopedia, much of the benefit of the generator-driven paradigm can be achieved using the source-rephrasing paradigm and a simple summarization prompt.

\item \textbf{Increasing token information density improves model performance}: The fact that naive summary prompts achieve a \textcolor{blue}{+1.2pp} improvement over the baseline (45.5\%) indicates that increasing per-token information density is one mechanism (among potentially many) by which synthetic data improves pre-training.
\item \textbf{Synthetic data is not just knowledge distillation}: While summarization shows similar benefits as Cosmopedia, \beyondweb outperforms summarization-based approaches with 50.4\% accuracy (\textcolor{blue}{+3.7pp}). This highlights that substantial gains are possible beyond basic distillation and underscores the importance of careful synthetic data generation.

\end{enumerate}

\begin{tcolorbox}[colback=beyondwebbluelight, colframe=beyondwebbluedark, coltitle=white, title=Takeaway: Synthetic Data Is Not Just Knowledge Distillation]
We find that a simple summarization-based rephrasing approach is able to nearly match the performance of the much more expensive generator-driven approach of Cosmopedia. This highlights that increasing per-token information density is one mechanism (among potentially many) by which synthetic data improves pre-training.
However, \beyondweb outperforms summarization-based approaches by a clear margin, showing that substantial gains are possible beyond basic distillation, and underscoring the importance of intentional synthetic data approaches.
\end{tcolorbox}

\subsection{RQ2: Can Synthetic Data Break the Data Wall?}
\label{sec:data_wall}

Can synthetic data effectively compensate for limited high-quality real-world data? This section explores whether there exists a fundamental limit to model performance that cannot be overcome through synthetic data. Our results reveal a surprising finding: \textbf{the data wall is surpassable \textit{depending on the type of synthetic data}}.

\begin{figure}[!ht]
    \centering
    \includegraphics[width=0.8\textwidth]{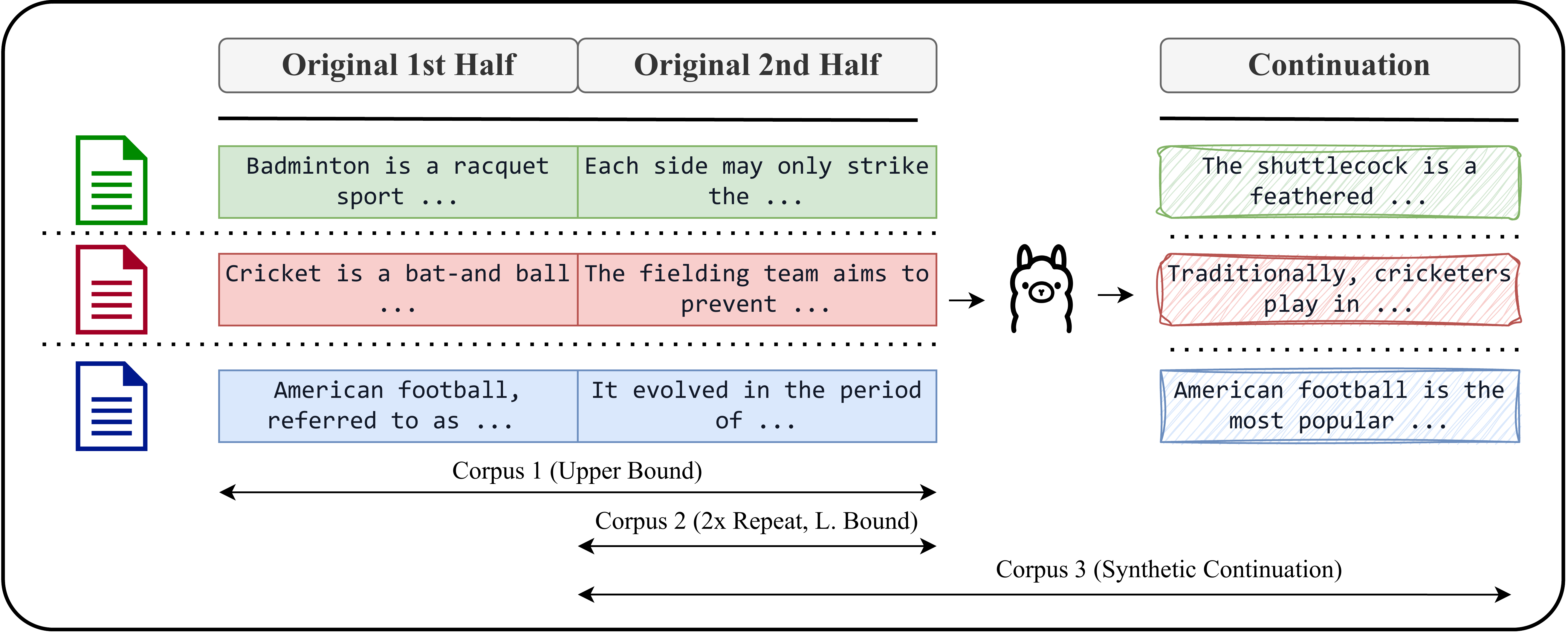}
    \caption{
        Illustration of data splitting and corpus construction strategies to enable a controlled setup. The figure shows how our 20 billion token dataset is divided and utilized across different experimental conditions. The top row displays three data segments: \textbf{Original 1st Half} (10B tokens of natural web content), \textbf{Original 2nd Half} (10B tokens of natural web content), and \textbf{Continuation} (10B tokens of synthetic content generated by extending documents from the first half). The example text snippets demonstrate how continuation generates stylistically consistent but novel content. The arrows below indicate corpus composition: \textbf{Corpus 1 (Upper Bound)} uses both original halves for full natural data coverage; \textbf{Corpus 2 (2x Repeat)} uses only the first half repeated twice; and \textbf{Corpus 3 (Synthetic Extension)} combines the second half with synthetic continuations. This experimental design isolates the effects of repetition versus synthetic augmentation when facing data constraints.
    }
    \label{fig:experimental_setup}
\end{figure}

\begin{tcolorbox}[title=Continuation Prompt]
\label{fig:summary_prompt}
Continue the following text in the same style as the original.
\end{tcolorbox}

\paragraph{Experiment Design.} We designed three controlled datasets (see Figure \ref{fig:experimental_setup}) to systematically compare different training approaches operating in a data-constrained setting of 20 billion tokens.

\begin{enumerate}[leftmargin=*]
\item \textbf{Full Data (Upper Bound).} This represents our performance ceiling, utilizing the complete available dataset of 20 billion tokens without any repetition or synthetic augmentation. This corpus serves as the \textit{oracle} baseline, representing the maximum achievable performance given our data constraint. The figure shows this as a complete, unbroken dataset.

\item \textbf{2x Repeat (Lower Bound).} This corpus simulates data scarcity by artificially constraining our dataset to only 10 billion unique tokens, then repeating this exact content twice to reach our full 20 billion token training budget. We split each paragraph at the sentence boundary closest to its midpoint, and only retain the second half of the paragraph. This approach represents a naive solution of repeating data to meet training token requirements, and serves as our lower bound.

\item \textbf{Continuation (Naive Synthetic).} Building on the same 10 billion token dataset as the previous strategy, this approach generates the remaining 10 billion tokens through simple model-driven continuation. An LLM (Llama-3.1-8B in this case) receives partial documents from the \emph{second half} of the original dataset and is prompted to generate natural continuations. This represents the most straightforward approach to synthetic data generation, where models extend the available data. The synthetic content attempts to maintain stylistic consistency with the source material.
We specifically chose the \emph{second} half of each paragraph in the original data to avoid the confounding effect of the generator model already having seen the full sample during its own pretraining. If the generator model had seen the sample, then it may be able to `cheat' by generating the true second half when prompted with the first half, and hence approximate the full data upperbound. This would allow us to test whether naive continuation of part of the data can approximate the whole.
\end{enumerate}

\begin{figure}[!ht]
    \centering
    \includegraphics[width=0.6\textwidth]{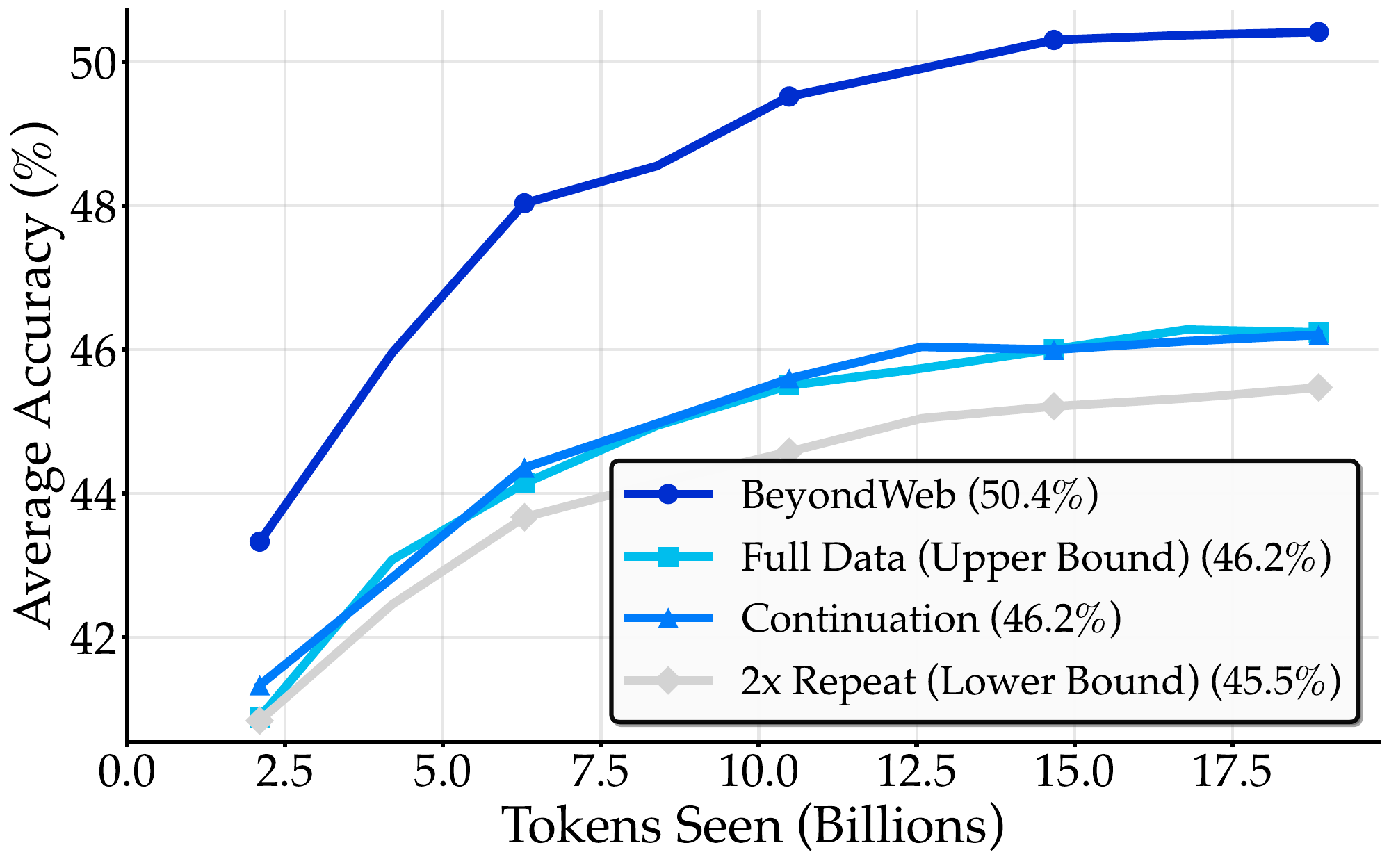}
    \caption{
        Performance comparisons across different data augmentation strategies during training. The \textcolor{beyondwebbluedark}{dark blue} line represents \textbf{\beyondweb (50.4\%)} which significantly surpasses all other approaches. The \textcolor{customlightblue}{light blue} line shows \textbf{Continuation (46.2\%)}, the \textcolor{cyan}{cyan} line depicts \textbf{Full Data Upper Bound (46.2\%)}, and the \textcolor{gray}{gray} line represents \textbf{2x Repeat Lower Bound (45.5\%)}. The striking visual separation emphasizes \beyondweb's \textcolor{blue}{+4.2pp} improvement over the Full Data upper bound. This reflects how intentionality is critical to breaking the data wall with synthetic data, and not just \emph{any} synthetic data will yield benefits.}
    \label{fig:data_wall_results}
\end{figure}

\paragraph{Results and Observations.}

Figure~\ref{fig:data_wall_results} offers a visual comparison of average accuracy across models pretrained on different datasets using various augmentation strategies.

\begin{enumerate}[leftmargin=*]
\item \textbf{Repetition leads to performance degradation}: The 2x Repeat strategy (gray line) achieves only 45.5\% accuracy, representing a 0.7pp drop from the full data baseline (46.2\%). This demonstrates that naive data duplication leads to diminishing returns as seen in prior works~\citep{muennighoff2023scaling,goyal2024scaling}.

\item \textbf{Naive synthetic generation provides only modest improvements}: The Continuation strategy (46.2\%) offers a small \textcolor{blue}{+0.7pp} gain over 2x Repeat (45.5\%). This matches the Full Data upper bound, suggesting that simple model-generated extensions can potentially compensate for limited quantities of web data. We note that an important confounder in this experiment is that the continuation model has seen a lot more than the 20B tokens available for full data experiment. This means that it may be using its parametric knowledge to add `new knowledge'.

\item \textbf{Strategic synthetic data breaches the data wall}: \beyondweb achieves 50.4\% accuracy, surpassing the Full Data upper bound (46.2\%) by \textcolor{blue}{+4.2pp}. This finding demonstrates that carefully crafted synthetic data can exceed the performance ceiling of natural data. The data wall is not unsurpassable; it can be broken through strategic synthetic data generation.

\end{enumerate}

\begin{tcolorbox}[colback=beyondwebbluelight, colframe=beyondwebbluedark, coltitle=white, title=Takeaway: Synthetic Data Generation Must be Thoughtful to Improve over Real Data]
Simple continuation provides limited improvement over repetition, and may not be able to breach the data wall. Synthetic data must be well designed, as shown by \beyondweb, which can significantly exceed the performance of training solely on natural data.
\end{tcolorbox}

\subsection{RQ3: How Important is the Quality of the Seed Data for Rephrasing?}
\label{sec:quality_matters}

An important design decision in synthetic data generation involves the quality of both the source material and the generated synthetic content. This raises a fundamental question: \emph{if one has limited high-quality data, is it better to use high-quality (but repeated) input data to seed synthetic rephrasing, or leverage lower-quality sources and turn them into high quality?}

\paragraph{Experiment Design.}
We operate under a constraint of 10B tokens of high-quality data, along with an abundant supply of low-quality data.
Following the method from \citet{datologyai_text_2024}, we sample a small subset of high-quality (HQ 
Web) data from RedPajama. We also use a random sample of the RPJ dataset as our low-quality subset (LQ Web). We then rephrase the LQ Web data to yield LQ Synth and the HQ Web data to yield HQ Synth, and compare training scenarios that use different quality combinations of synthetic and/or web data. We note that we rephrase the \emph{same} HQ web data that we also use in the data mix.

\begin{figure}[!ht]
    \centering
    \includegraphics[width=0.6\textwidth]{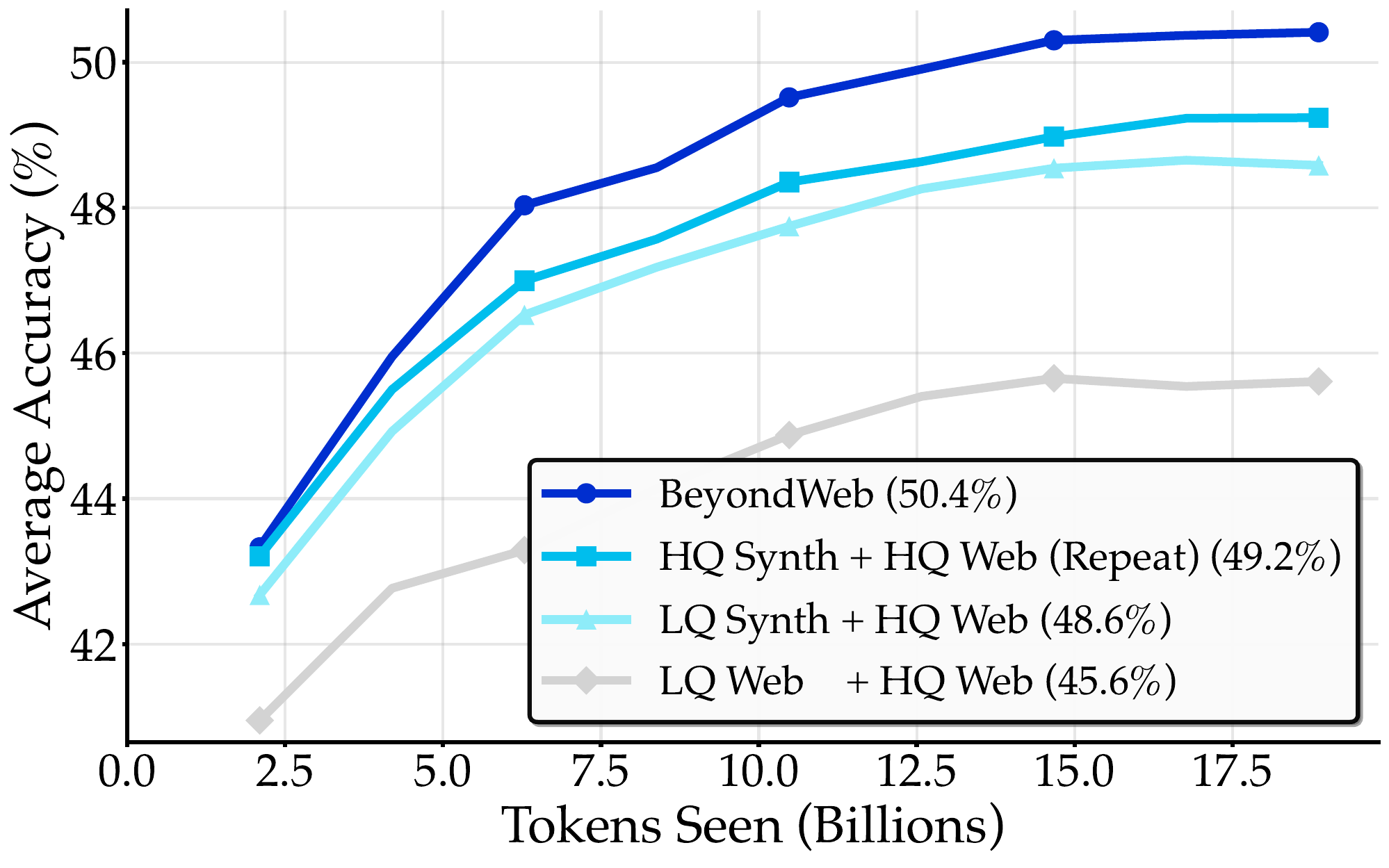}
    \caption{
        Performance comparison across different quality combinations in training data. \textbf{HQ} refers to high-quality web data, LQ refers to low-quality web data. The \textcolor{beyondwebbluedark}{dark blue} line shows \beyondweb(50.4\%), \textcolor{cyan!100}{dark cyan} shows \textbf{HQ Synth + HQ Web (49.2\%)}, where the synthetic data are rephrased versions of the HQ web samples, and the \textcolor{cyan!50}{light cyan} line shows \textbf{LQ Synth + HQ Web (48.6\%)}. The \textcolor{gray}{gray} baseline corresponds to \textbf{LQ Web + HQ Web (45.6\%)}. These results indicate that improving the quality of input data for rephrasing improves the rephrased data, even when there is overlap with the original input data. But improved input data quality alone is inadequate for producing the highest quality synthetic data. 
    }

    \label{fig:quality_matters}
\end{figure}

\paragraph{Results and Observations.}

Figure \ref{fig:quality_matters} shows two important insights about input data for rephrasing:

\begin{enumerate}[leftmargin=*]
\item \textbf{High-Quality Seed Data Dominates}: The high performance of the combination of HQ Synth + HQ Web (49.2\%) demonstrates that starting with high-quality source material for synthetic rephrasing is crucial. Notably, the \beyondweb performance at 50.4\% shows that there remain other avenues for performance improvement beyond high-quality source data.

% \item \textbf{Quality Trumps Novelty}: Although the LQ Synth + HQ Web combination (48.6\%) outperforms LQ Web + HQ Web, consistent with the findings of \citet{nguyen2025recyclingwebmethodenhance}, it still lags behind HQ~Synth + HQ~Web (49.2\%). This highlights that the quality of seed data for synthetic generation matters more than ensuring complete novelty in the knowledge base.

\item \textbf{Quality Trumps Novelty}: The LQ Synth + HQ Web combination (48.6\%) underperforms the HQ Synth + HQ Web data (49.2\%), indicating that the quality of seed data for synthetic generation matters more than ensuring complete novelty in the knowledge base. However, it still provides a \textcolor{blue}{+3.0pp} improvement over the baseline, consistent with the findings of \citet{nguyen2025recyclingwebmethodenhance}.

\end{enumerate}

\begin{tcolorbox}[colback=beyondwebbluelight, colframe=beyondwebbluedark, coltitle=white, title=Takeaway: Quality Synthesis Can Enable Knowledge Amplification]
Given limited high-quality data, it is more beneficial to use it as seed data for synthetic rephrasing rather than low-quality data, despite the possible drawback of repeating knowledge from the high-quality sources. Nevertheless, high-quality input data alone is insufficient for generating the highest quality synthetic data.
\end{tcolorbox}

\subsection{RQ4: Is Distributional Style Matching Important?}
\label{sec:style_matters}

While vast amounts of text data are available online, the natural distribution of web content is different from the types of text typically encountered at inference. For instance, the internet is dominated by personal blogs, news articles, and product pages~\citep{wettig2025organize}. However, evaluation use cases typically involve chat, or in-context learning abilities. These can together be stylistically seen as a `ping-pong' rally between questions and answers, or users and assistants. We investigate the impact of this implicit distribution shift between train and test, and if such stylistic alignment can be one mechanism by which synthetic rephrasing can improve model performance.

\paragraph{Estimating the amount of conversational web data.}
We first estimate the fraction of conversational data in the general web corpus. To do so, we sampled 10k random examples from the RedPajama dataset, and queried the \texttt{gpt4o} model to label them as conversational or not. Here conversational refers to various forms of texts which have a `ping-pong' effect (see Appendix~\ref{sec:quantify_conv} for prompt). Our analysis indicated that conversational dialogue comprises less than 2.7\% of internet text. Yet, the dominant use case for modern language models lies in chat-based applications, such as virtual assistants, customer service bots, and interactive tools.

This discrepancy reveals a gap between the training distribution and real-world model usage. The formats and interaction styles most relevant to deployment, such as multi-turn dialogue, question answering, and instructional content, are significantly underrepresented in naturally occurring data. Synthetic rephrasing provides a mechanism to bridge this gap by generating examples in the specific forms and contexts that matter for downstream applications.

\paragraph{Estimating the impact of conversational web data.}

To systematically evaluate the impact of style alignment, we conduct controlled experiments that vary the proportion of conversational content in our training data while maintaining constant token budgets and data quality.
We leverage the style classification filters from Organize the Web~\citep{wettig2025organize} to identify naturally occurring conversational content within the RedPajama dataset. From their 20 identified web content styles, we focus on four that exhibit conversational characteristics: \texttt{Audio Transcript, Customer Support, FAQ, and Q\&A Forum}. Manual inspection confirmed these categories contain the back-and-forth dialogue patterns characteristic of conversational interaction.

As a validation of our previous investigation, this naturally occurring conversational content comprises 3.67\% of the RedPajama dataset, roughly aligning with our GPT-based estimate of 2.7\% mentioned earlier. We construct three training datasets with varying conversational ratios: 10\%, 20\%, and 50\% conversational content, with the remainder consisting of randomly sampled RedPajama data. Each dataset maintains our standard 20 billion token training budget, enabling direct performance comparisons across different style distributions.

\begin{figure}[!ht]
    \centering
    \includegraphics[width=0.55\textwidth]{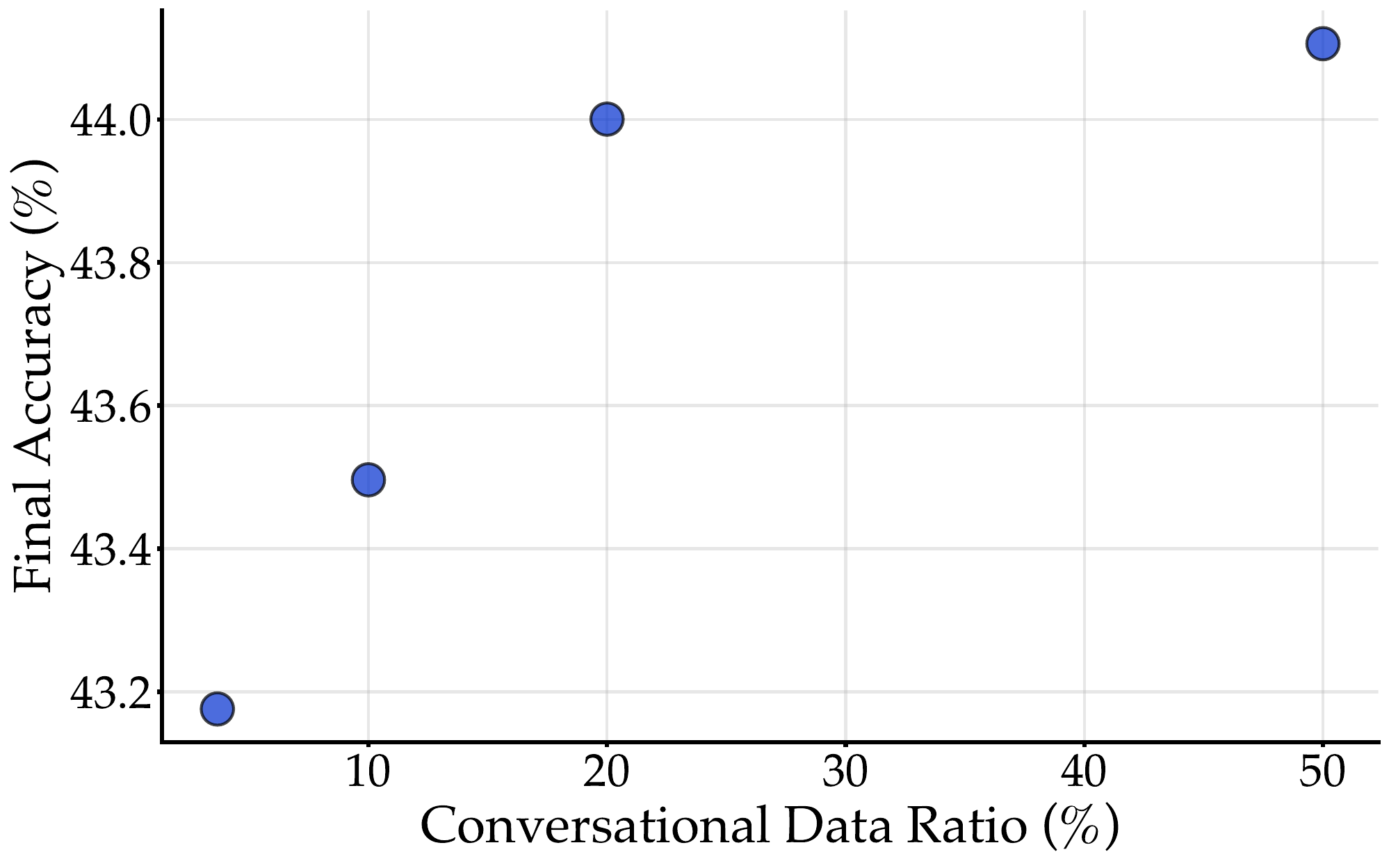}
    \caption{\textbf{Effect of conversational data ratio on final accuracy.}  
    The first point corresponds to the \textbf{RPJ baseline} (randomly sampled) which contains \textbf{3.67\% conversational data}.  
    For each of the other points, we \textbf{upsampled conversational data} in RPJ to the desired ratio and replaced the remaining portion with randomly sampled RPJ data to achieve the 20B token training budget. Increasing the percentage of conversational data beyond the 3.67\% baseline can improve performance by up to 0.9pp at 50\% conversational data, but the gains saturate beyond 20\%, indicating that style-matching is important, but not sufficient for maximizing synthetic data quality.}
    \label{fig:conversational_scatter_only}
\end{figure}

\paragraph{Results and Observations.}
Since our focus is on multi-turn, back-and-forth conversations, for this evaluation we measure only the 5-shot performance of the trained models.
We find that increasing the proportion of conversational content in pretraining improves model performance (Figure~\ref{fig:conversational_scatter_only}). 
The first point represents the \textbf{RPJ baseline}, which contains 3.67\% conversational data from random sampling. 
For higher ratios, we upsample conversational data within RPJ to the target percentage and replace the remaining portion with randomly sampled RPJ data. 
Performance increases from 43.2\% at the baseline to 44.1\% at 50\% conversational content, with 10\% and 20\% conversational data yielding 43.5\% and 44.0\%, respectively. 
This trend confirms that increased exposure to conversational patterns during pretraining modestly enhances downstream task capabilities, though gains exhibit diminishing returns.

\begin{tcolorbox}[colback=beyondwebbluelight, colframe=beyondwebbluedark, coltitle=white, title=Takeaway: Distributional Style Matching is Useful but not Sufficient]
Conversational content is a small fraction of web data (3.67\%), yet chat and in-context learning are the primary inference use cases of LLMs. Upsampling natural conversational data improves downstream task performance, though these improvements are modest and show diminishing returns with the proportion of conversational data. These results indicate that style-matching
can improve synthetic data quality, but is not sufficient for maximizing it.
\end{tcolorbox}

\subsection{RQ5: How Important is Diversity at Scale?}
\label{sec:diversity}

In the previous section, we found that aligning styles between train and test time has positive but diminishing gains. This finding highlights a critical limitation: while style alignment helps initially, models eventually saturate on uniform synthetic data patterns. This leads us to examine the importance of diversity in synthetic data generation.
We investigate a fundamental question: How does synthetic data diversity affect model performance when scaling to trillions of tokens? Specifically, do diverse generation strategies maintain benefits throughout extended training, or do all approaches eventually plateau?
Our experiments reveal that different synthetic data generation approaches yield dramatically different scaling outcomes, with approaches that emphasize diversity, exemplified by \beyondweb, significantly outperforming fixed synthetic data generation techniques.

\begin{figure}[!ht]
    \centering
    \begin{minipage}{0.32\textwidth}
        \centering
        \includegraphics[width=\textwidth]{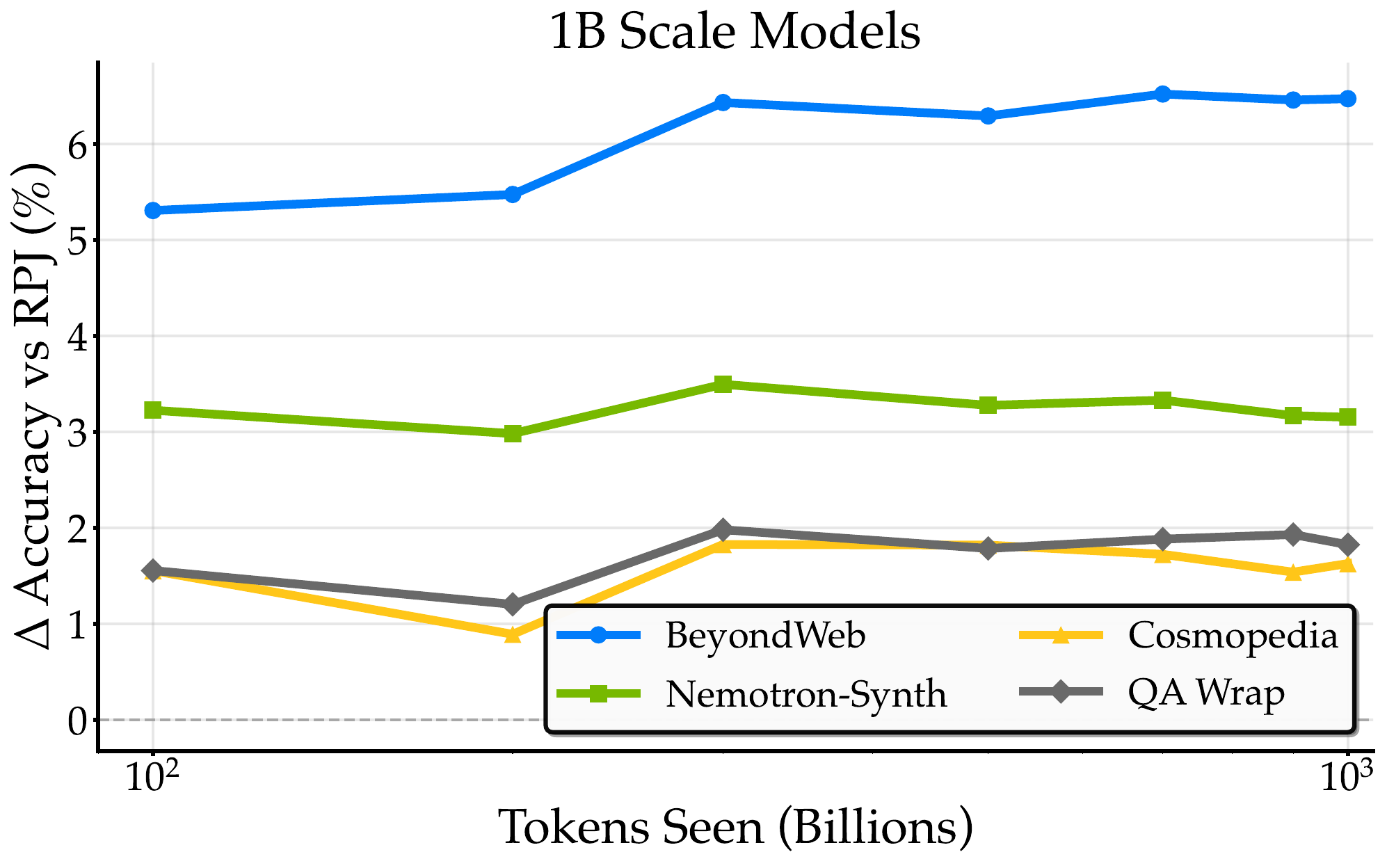}
    \end{minipage}
    \hfill
    \begin{minipage}{0.32\textwidth}
        \centering
        \includegraphics[width=\textwidth]{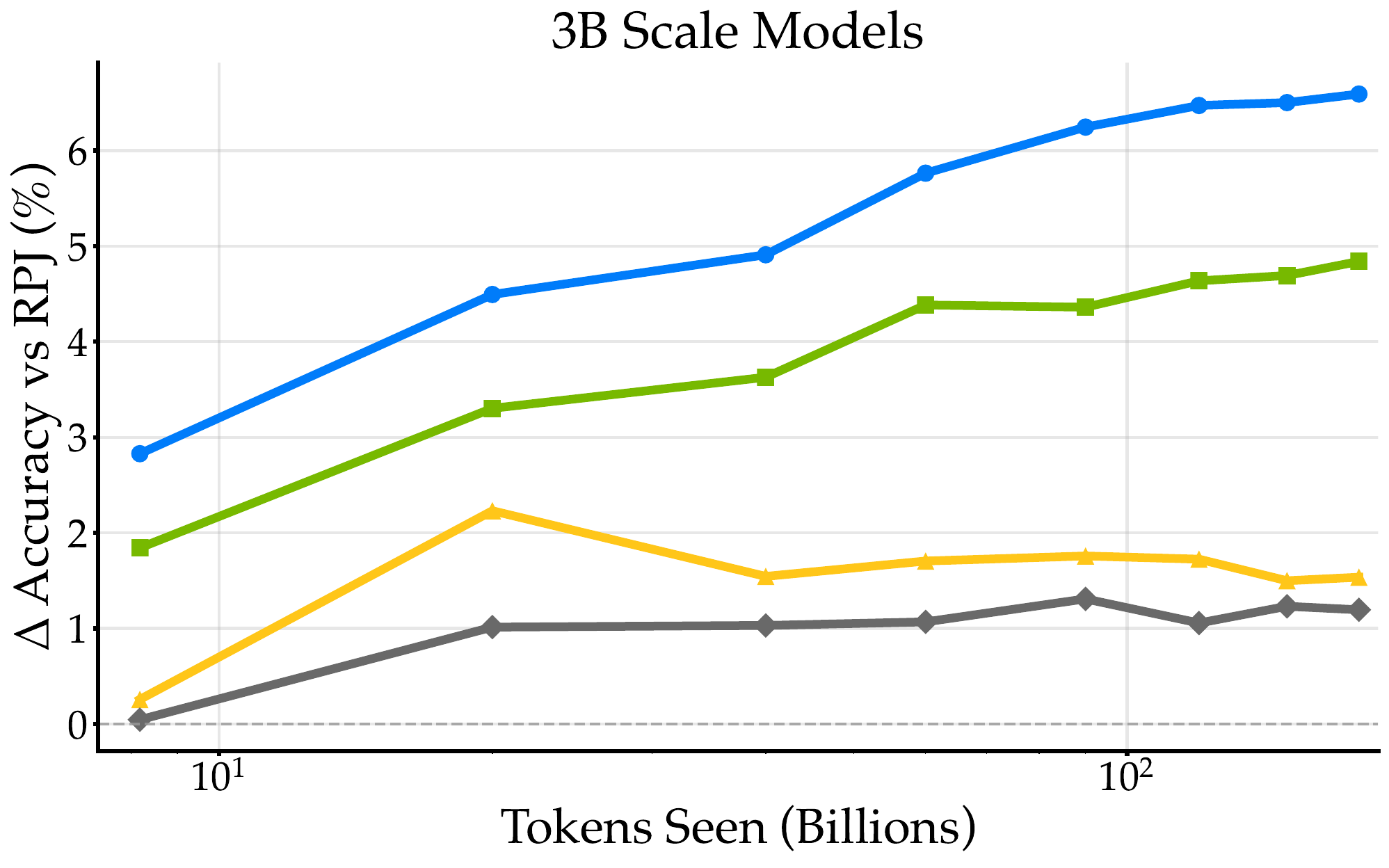}
    \end{minipage}
    \hfill
    \begin{minipage}{0.32\textwidth}
        \centering
        \includegraphics[width=\textwidth]{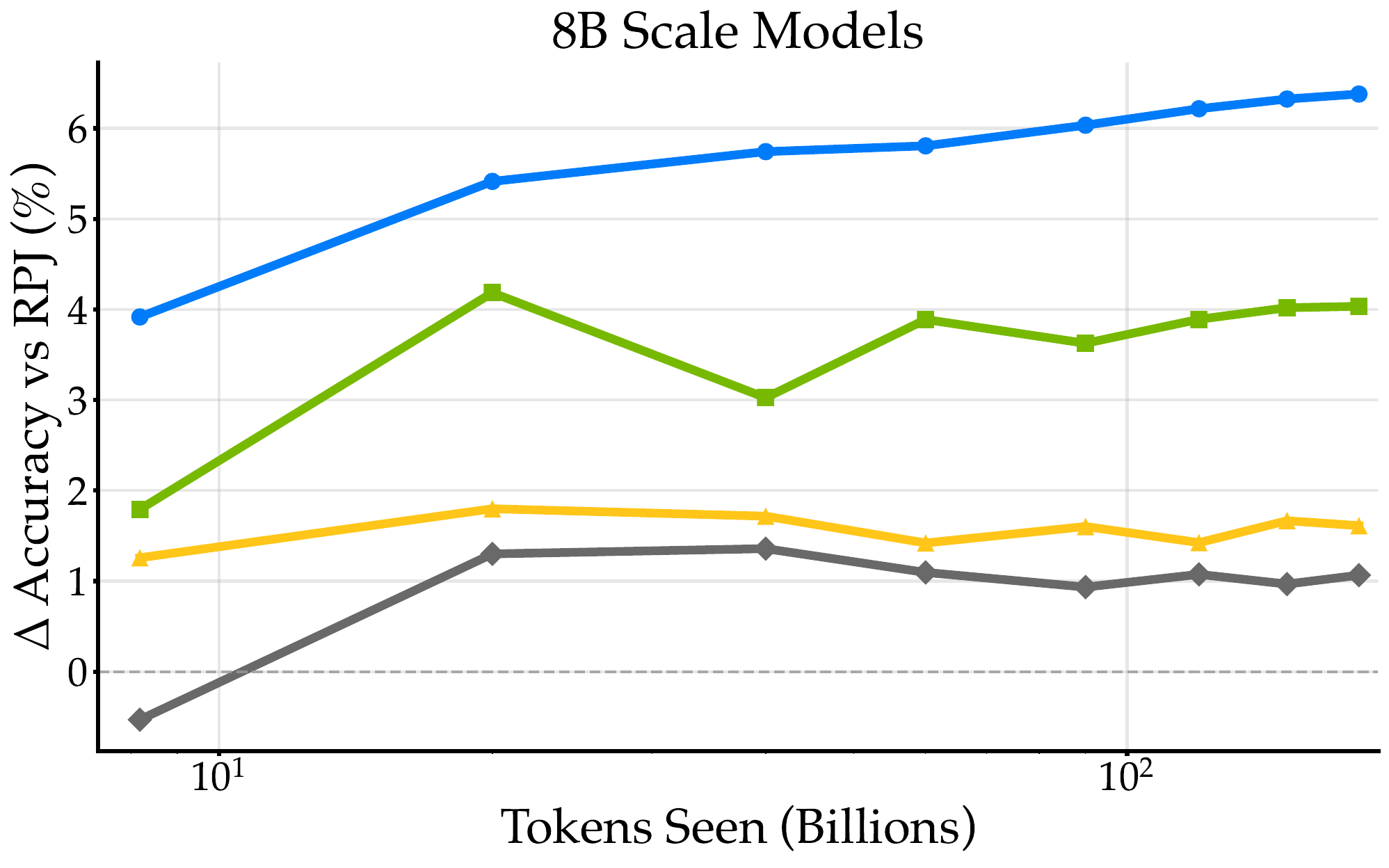}
    \end{minipage}
    \caption{Training dynamics across model scales showing performance improvements relative to RedPajama baseline. \textbf{Left:} 1B Scale, \textbf{Center:} 3B Scale, \textbf{Right:} 8B Scale. \beyondweb (blue) maintains consistent positive slopes across all scales, while QA approaches (cyan) show early gains followed by plateauing effects, particularly evident in larger models. Notably, at 1B scale, models are trained to saturation at approximately 50$\times$ beyond Chinchilla optimal compute. Where other baselines begin to overfit and degrade, \beyondweb continues to grow in its improvements over RedPajama performance. The sustainable improvement curves demonstrate that diverse generation strategies provide more robust long-term benefits than single-strategy approaches across training regimes. Note that in order to smooth intermediate fluctuations, each data point on the plot is the average accuracy of multiple previous intermediate checkpoints (e.g. the 100B token checkpoint is the average of all checkpoints until 100B, the 200B checkpoint is the average of all checkpoints between 100B and 200B, etc).}

    \label{fig:hero_runs_combined}
\end{figure}

\paragraph{Experiment Design.}

To understand the scaling effects of different synthetic data strategies, we analyze training dynamics across model scales through extended training runs. Among the strategies we investigate, Cosmopedia generates a textbook related to a web sample under the generator-driven paradigm, QA Wrap rephrases existing web documents into conversational question-answers. Nemotron-Synth builds on top of WRAP by diversifying the prompts beyond just conversational question-answering, into various styles like MCQ, Yes/No questions, open-ended questions, reading comprehension tasks, logical problem solving tasks, and so on. \beyondweb further prioritizes diversity in synthetic data generation, and prioritizes learning patterns that aid learning and information retention.

\paragraph{Results and Observations.}

The training dynamics reveal fundamental differences between generation strategies as we scale to hundreds of billions of tokens of data, particularly evident across different training regimes. Diversity in synthetic data generation manifests in two critical ways:

\begin{enumerate}[leftmargin=*]
\item \textbf{Diversity provides an initial performance bump:} Multi-strategy approaches like \beyondweb provide substantial early gains as models quickly benefit from 
immediate exposure to
diverse formats and styles. We hypothesize this aids learning by making patterns that align with model's final usage, core to its early representations.

\item \textbf{Diversity leads to sustained improvements:} Diverse generation strategies continue providing learning benefits throughout extended training, even in overtrained regimes. At the 8B scale, the marginal improvements for \beyondweb over RPJ (y-axis) continue to increase, whereas the benefits for strategies such as Cosmopedia (which generates a fixed style of textbook content) tend to saturate (Figure~\ref{fig:hero_runs_combined}). Remarkably, at 1B scale where models are trained approximately 50$\times$ beyond Chinchilla optimal compute, \beyondweb maintains positive performance trends while other baselines begin to overfit and show degraded performance, demonstrating the robustness of diverse synthetic data against overfitting in extreme training regimes.

\end{enumerate}

\begin{tcolorbox}[colback=beyondwebbluelight, colframe=beyondwebbluedark, coltitle=white, title=Takeaway: Diversity Enables Sustained Learning]
While standard synthetic data generation methods provide initial benefits as models adapt to consistent formats, their lack of stylistic diversity leads to diminishing returns.  Multi-faceted approaches continue providing valuable learning signals throughout training, whereas single-strategy methods saturate.
\end{tcolorbox}

\subsection{RQ6: How Important is the Synthetic Rephraser Model Family?}
\label{sec:family_consistency}

A key difference between the source rephrasing and the generator-driven approach is that the synthetic data generator is not required to be the `source' of the knowledge. Instead it only needs to transform the data to make it more useful for pretraining. While existing work like Cosmopedia \citep{benallal2024cosmopedia} and Phi-1.5~\citep{li2023textbooks} requires large generator models, Mixtral-8x7B-Instruct-v0.1~\citep{jiang2024mixtralexperts} and GPT-4, respectively, we question whether the specific generator family used for rephrasing significantly impacts the quality of the rephrased data.

\paragraph{Experiment Design.}

We test the importance of generator family by using 4 different generator models: (i) OLMo-2-7B\footnote{\url{https://huggingface.co/allenai/OLMo-2-1124-7B-Instruct}} \citep{olmo20242olmo2furious}, (ii) Phi-4-14B\footnote{\url{https://huggingface.co/microsoft/phi-4}} \citep{abdin2024phi4technicalreport}, (iii) Mistral-7B-v0.3\footnote{\url{https://huggingface.co/mistralai/Mistral-7B-Instruct-v0.3}} \citep{jiang2023mistral7b}, and (iv) Llama-3.1-8B\footnote{\url{https://huggingface.co/meta-llama/Llama-3.1-8B-Instruct}} \citep{grattafiori2024llama}. We use the same rephrasing prompt for each model and evaluate the performance of a model trained on the synthetic data generated by each of these generator models. We generated 10B tokens of synthetic data using each of them, combined it with 10B tokens of web data, and trained a model on each of the resulting 20B token datasets.

\paragraph{Results and Observations.}
Synthetic data rephrased using different model families consistently produces high-quality training data, with the resulting datasets achieving gains ranging from \textcolor{blue}{+3.4pp} to \textcolor{blue}{+4.4pp} over the RPJ baseline (45.5\%). All model families were within \textcolor{blue}{±1pp} of each other, with OLMo-2-7B achieving the highest performance (49.9\%) and Mistral-7B-v0.3 the lowest (48.9\%).

We compare the general benchmark performance of each rephraser model (OLMo-2-7B: 59.6\%, Llama-3.1-8B: 61.2\%, Mistral-7B-v0.3: 66.0\%, and Phi-4: 66.6\%) with the quality of the synthetic data they produce, measured by the performance of models trained on their rephrased outputs. While the rephrasers span a 7-point range in general accuracy, the quality of the resulting synthetic datasets is strikingly similar, differing by less than 1 percentage point across generators. Notably, OLMo-2-7B, despite having the lowest benchmark accuracy, produces the highest-quality synthetic data in our evaluation (49.9\%). This lack of positive correlation indicates that a model's general language modeling ability does not determine synthetic data quality, highlighting that even comparatively weaker models can be highly effective rephrasers.

These findings suggest that the primary driver of performance is not the generator's general language modeling capabilities, but rather the limited and simpler capability to perform effective rephrasing and data transformation, which appears to be consistent across different model families. This generator-agnostic behavior means that organizations can deploy synthetic data pipelines using available model families without worrying about generator-specific optimizations, thereby reducing overall resource usage and enabling the viability of open-source and permissible models.

\begin{figure}[!ht]
    \centering
    \begin{minipage}{0.48\textwidth}
        \centering
        \includegraphics[width=\textwidth]{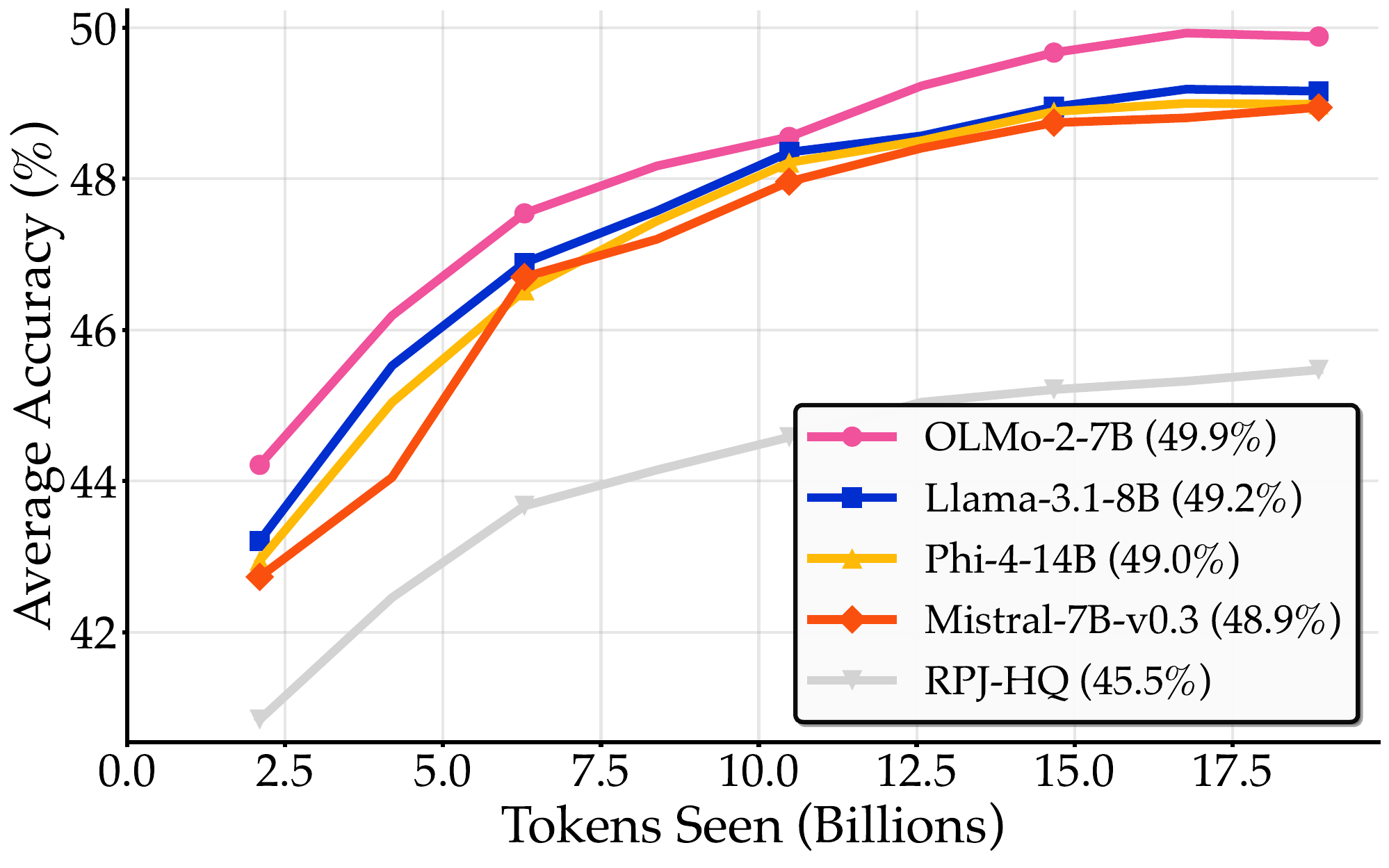}
    \end{minipage}
    \hfill
    \begin{minipage}{0.48\textwidth}
        \centering
        \includegraphics[width=\textwidth]{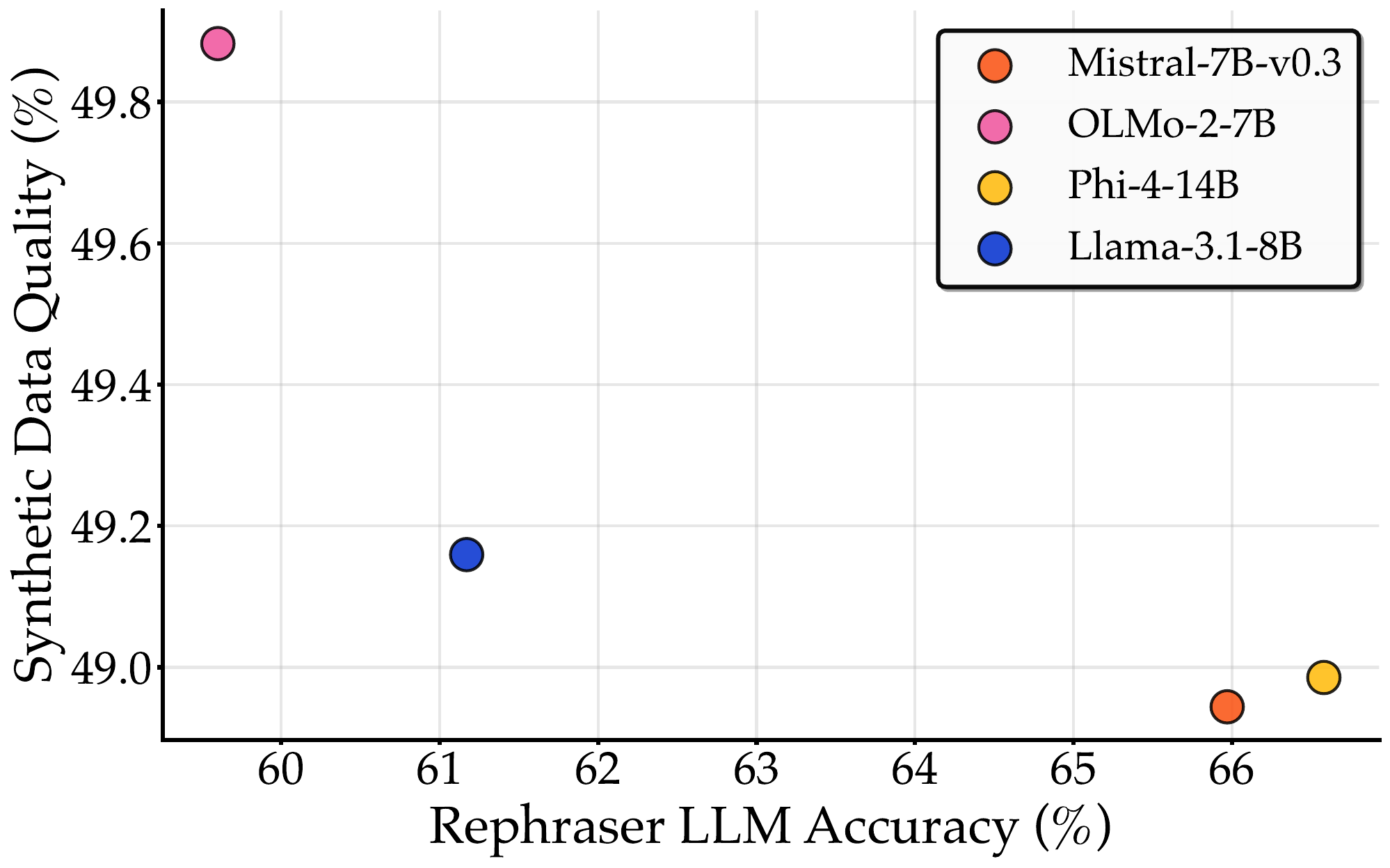}
    \end{minipage}
    \caption{Synthetic data benefits are largely consistent across generator model families, and generator model quality does not predict synthetic data quality. \textbf{Left:} This plot shows the quality of models trained on synthetic data generated by different models: the pink line represents data generated by OLMo-2-7B (49.9\%), the yellow line shows data generated by Phi-4-14B (49.5\%), the orange line indicates data generated by Mistral-7B-v0.3 (48.9\%), and the blue line depicts data generated by Llama-3.1-8B (49.2\%). Models trained on each of these synthetic datasets achieve substantial improvements over the gray baseline representing RPJ (45.5\%), with three out of the four models generating synthetic data of nearly identical quality, and OLMo-2-7B generating better data. \textbf{Right:} We plot the benchmark performance (on the x-axis) of the rephraser models used for synthetic data generation and contrast with the performance of the LLM trained on their generated data (marked as synthetic data quality on the y-axis). We observe that the benchmark performance of a language model does not positively correlate with its rephrasing capability.}
    \label{fig:family_comparison}
\end{figure}

\begin{tcolorbox}[colback=beyondwebbluelight, colframe=beyondwebbluedark, coltitle=white, title=Takeaway: Generator Knowledge is Less Important]
Synthetic data benefits are largely robust across generator families, with consistent improvements across Phi-4-14B, Mistral-7B-v0.3, and Llama-3.1-8B, and further improved synthetic data quality from OLMo-2-7B. Additionally, generator model quality is not predictive of synthetic data quality. These results indicate that selecting a \emph{good} rephrasing model may be straightforward as most work well, but selecting the \emph{best} rephrasing model is not.
\end{tcolorbox}

\subsection{RQ7: Does Rephraser Size Matter?}
\label{sec:generator_scaling}

Having observed near-invariance in the performance of synthetic data generated by different model families, we now turn to the question of whether the size of the rephraser model affects synthetic data quality. Rephrasing is a much more constrained task than open-ended generation, so it's possible that even small models may be able to generate high-quality synthetic data via rephrasing. This is once again in contrast to the generator-driven paradigm where the generator model is required to be the \emph{source} of the knowledge, and the quality of the synthetic data is heavily dependent on the size of the generator model~\citep{benallal2024cosmopedia}. 

\paragraph{Experiment Design.} To investigate this, we used Llama-3 models of varying sizes, 1B, 3B, and 8B, to generate 10B tokens of synthetic data. We combined the 10B synthetic tokens with 10B tokens of web data, and trained a model on each of the resulting 20B token datasets.

\begin{figure}[!ht]
\centering
\includegraphics[width=0.6\textwidth]{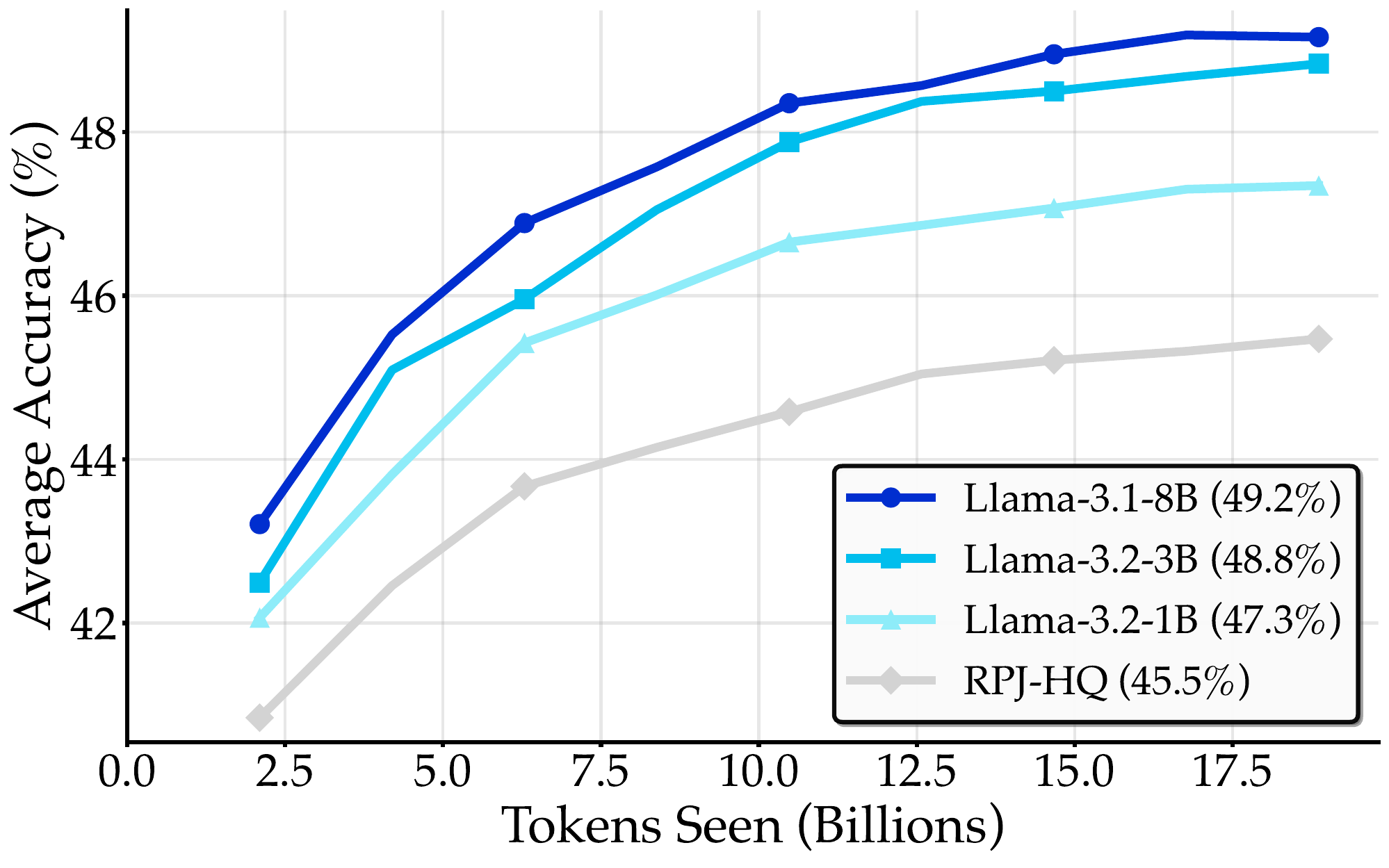}
\caption{Effect of generator model size on synthetic data quality. The plot demonstrates the impact of generator size: the light cyan line shows Llama-3.2-1B generator performance (47.3\%), the cyan line represents Llama-3.2-3B generator results (48.8\%), and the dark blue line indicates Llama-3.1-8B generator performance (49.2\%). The improvement from 1B to 3B (\textcolor{blue}{+1.5pp}) is substantial, while the gain from 3B to 8B (\textcolor{blue}{+0.4pp}) shows diminishing returns. The gray baseline at 45.5\% represents RPJ performance, emphasizing the consistent improvements achieved by all generator sizes.}
\label{fig:generator_size_comparison}
\end{figure}

\paragraph{Results and Observations.}
As depicted in Figure \ref{fig:generator_size_comparison}, larger generators consistently produce better synthetic data, yielding substantial downstream performance gains (\textcolor{blue}{+3.7pp} for 8B, \textcolor{blue}{+3.3pp} for 3B) over RPJ baseline (45.5\%). The 1B generator also shows significant benefit (\textcolor{blue}{+1.8pp}), demonstrating that small generators can be effective. Larger LLMs provide only small additive gains.

While the 8B model (49.2\%) slightly outperforms the 3B model (48.8\%), the small gain (\textcolor{blue}{+0.4pp}) indicates diminishing returns beyond 3B. This near-saturation makes small models a cost-efficient and practical choice for many applications, offering a strong balance between quality, speed, and compute overhead.
Larger generators likely offer advantages in knowledge depth, reasoning, and content diversity, which smaller models cannot replicate. However, this 3B capability threshold makes high-quality synthetic data accessible to researchers with limited computational resources. This observation, however, is specific to our setup and may vary under more sophisticated generation methods or on more demanding benchmarks.

\begin{tcolorbox}[colback=beyondwebbluelight, colframe=beyondwebbluedark, coltitle=white, title=Takeaway: Small Models Can Produce High Quality Synthetic Data]
The quality of synthetic data increases when increasing generator size from 1B to 3B, then starts to saturate at 8B. The simplicity of rephrasing makes generator size less critical, enabling highly scalable synthetic pretraining data generation even with small models.
\end{tcolorbox}

%% file: sections/5-future.tex
\section{Future Directions}

Our work with synthetic data opens several research directions that warrant further investigation:

\paragraph{Scaling Laws and Inherent Repetition.}

A crucial area for future research lies in understanding the scaling laws specific to synthetic data. Unlike real data, where repetition is explicit and measurable, synthetic data presents a more nuanced form of inherent repetition stemming from the generating model's architecture and the prompting methods used. This intrinsic repetition, governed by the model's parameters and rephrasing strategies, requires new theoretical frameworks and metrics for quantification.
The development of such metrics would enable us to better assess the true value and diversity of synthetic data. This understanding is particularly important as we scale up synthetic data generation, as it could reveal fundamental limits or opportunities that differ from those observed in traditional scaling laws for real data.

\paragraph{Democratizing Synthetic Data Generation.}

An important observation from our work is that the task of rephrasing, at its core, doesn't necessarily require extensive world knowledge. This suggests the possibility of using much smaller models for synthetic data generation, which could significantly reduce computational costs. Future work should explore the minimal model size required for effective rephrasing while maintaining quality, potentially making synthetic data generation more accessible to researchers with limited computational resources.

\paragraph{Alignment with Human Values.}
Since the synthetic data generation process offers greater control, it creates promising opportunities for developing pretraining methodologies that explicitly promote alignment with human values~\citep{korbak2023pretraining}. This aspect of synthetic data training could provide a powerful tool for addressing alignment challenges in language model development, offering an alternative to post-hoc alignment techniques~\citep{maini2025safety}.

\paragraph{Applicability Across Data Domains.} While the present work, like most documented applications of source rephrasing, uses web data as input, the approach is not inherently limited to web data or even the text modality. It can be applied to domain-specific or proprietary data without requiring the rephrasing model to possess domain expertise, making it broadly suitable across diverse settings. Future work should explore applying such techniques to new domains and modalities, using synthetic data generation to help overcome the inherent data wall.

These research directions highlight the potential of synthetic data to not only improve model performance but also to address fundamental challenges in machine learning, from accessibility and scalability to alignment and safety. Future work in these areas could significantly impact how we approach language model training and deployment. 

%% file: sections/Conclusion.tex
\section{Conclusion}
\label{sec:conclusion}

Pretraining now routinely encounters a data wall: the supply of high-quality, information-dense web text can not keep up with modern training budgets. Synthetic data has emerged as a natural and effective response, yet the scientific understanding of when and how it helps has remained limited. We address this by studying synthetic data for pretraining at scale and introduce \beyondweb, a rephrasing-driven approach grounded in real web documents and diversified across styles and formats.

Our work identifies three general principles for effective synthetic data generation: prioritize quality over novelty by rephrasing high-quality sources rather than adding low-quality content; align the training data style with the deployment use-case; and maintain diversity in generation to sustain improvement over long training horizons. We show that these benefits are robust across generator families and saturate beyond moderate rephraser sizes, indicating that effective synthetic data generation does not require a particular model architecture or very large generators. However, we also show that the contribution of each of these factors in isolation is often modest, and no single factor is sufficient to yield high-quality synthetic data.

Taken together, these results demonstrate that high quality synthetic pretraining data is challenging to generate, even more challenging to ensure the efficacy of for larger models and training budgets, and risks being extremely costly to generate. Furthermore, there is no silver bullet for synthetic data, strong outcomes require jointly optimizing many variables. However, all of these challenges and risks are surmountable with thoughtful, scientifically rigorous source rephrasing, as evidenced by \beyondweb substantially outperforming other public synthetic pretraining data offerings.

\section{Contributions and Acknowledgements}
\label{sec:contri}

\begin{tabularx}{\textwidth}{@{}p{0.19\textwidth}X@{}}
\textbf{Core Research} & Pratyush Maini and Vineeth Dorna. \\[0.25em]
& \emph{steering the ship, wrangling the data, and keeping the Celsius industry thriving.} \\
\noalign{\vspace{0.75em}}

\multirow{2}{*}{\shortstack[l]{\textbf{Core Infra}}}  
& Parth Doshi, Aldo Carranza, Fan Pan, Jack Urbanek and Paul Burstein. \\
\noalign{\vspace{0.25em}}
& \emph{the all-star cast who turned Mazra into code, numbers, and wins.} \\
\noalign{\vspace{0.75em}}

\textbf{Leadership} & Bogdan Gaza, Ari Morcos, and Matthew Leavitt. \\[0.25em]
& \emph{the wise captains that made sure we didn’t steer into a bad baseline.} \\
\noalign{\vspace{0.75em}}

\textbf{Contributors} & Alex Fang, Alvin Deng, Amro Abbas, Brett Larsen, Cody Blakeney, Charvi Bannur, Christina Baek, Darren Teh, David Schwab, Haakon Mongstad, Haoli Yin, Josh Wills, Kaleigh Mentzer, Luke Merrick, Ricardo Monti, Rishabh Adiga, Siddharth Joshi, Spandan Das, and Zhengping Wang. \\
\noalign{\vspace{0.25em}}
& \emph{the ping-pong maestros who kept latency low and rallies long.}\\
\noalign{\vspace{0.75em}}

\textbf{Acknowledgements} &
Gem De Leon and Kristin Reinke for fueling us with Beyond Burgers and unshakable good vibes. Jacqueline Liu and Tiffanie Pham for assembling the all-star cast that made this work possible. Liz Gatapia for the beautiful logo design.
Kylie Clement, Jeremy Custenborder and Elise Clark for their sharp-eyed feedback on the draft.\\
\noalign{\vspace{0.25em}} 
& \emph{the backstage crew who kept us fed, staffed, and focused.}
\end{tabularx}

%% file: sections/appendix.tex
\section{Appendix}

\subsection{Additional Training Details}
\label{sec:app_training}

For all trainings, we use AdamW with $\beta_1{=}0.9$, $\beta_2{=}0.95$, a learning rate of $5e^{-4}$, and a weight decay of $1e^{-7}$. A linear warmup schedule was applied with 4K steps for the 1B model and 16K steps for the 3B and 8B models. Training was performed using fully sharded data parallelism (FSDP), with a batch size of 512, and a context length of 2048. 

\subsection{Evaluation Tasks}
\label{sec:eval_tasks}
We evaluate using lighteval \citep{lighteval} on a suite of 14 evaluation datasets that includes the 8 used by FineWeb \citep{penedo2024finewebdatasetsdecantingweb} and additional datasets that are well-established and demonstrate above-chance performance and monotonic improvement during training.

\begin{itemize}
    \item \textbf{ARC-Challenge} \citep{Clark2018ARC}: A dataset of 2,590 genuine grade-school level, multiple-choice science questions, designed to promote research in advanced question-answering. It challenges AI capabilities by requiring reasoning and comprehension beyond standard retrieval methods.
    
    \item \textbf{ARC-Easy} \citep{Clark2018ARC}: Similar to ARC-Challenge, but consists of 7,787 easier questions.
    
    \item \textbf{BoolQ} \citep{Clark2019BoolQ}: A question answering dataset for yes/no questions containing 15,942 examples. These questions are naturally occurring and generated in unprompted and unconstrained settings.
    
    \item \textbf{CommonsenseQA} \citep{talmor2019commonsenseqa}: A multiple-choice question answering dataset containing 12,102 questions requiring different types of commonsense knowledge.
    
    \item \textbf{COPA} \citep{roemmele2011copa}: Designed to assess commonsense causal reasoning through 1000 questions where each question consists of a premise and two alternatives.
    
    \item \textbf{HellaSwag} \citep{zellers2019hellaswag}: A benchmark dataset for commonsense natural language inference that challenges state-of-the-art models with context and endings that are easy for humans but difficult for machines.
    
    % \item \textbf{LAMBADA} \citep{paperno2016lambada}: Evaluates text understanding capabilities through word prediction tasks, focusing on managing long-term context and coherence.
    
    \item \textbf{MMLU} \citep{hendrycks2021mmlu}: A benchmark evaluating capabilities across 57 subjects, including mathematics, history, and law.
    
    \item \textbf{OpenbookQA} \citep{mihaylov2018openbookqa}: Contains 5,957 multiple-choice questions designed to probe understanding of core science facts and their applications.
    
    \item \textbf{PIQA} \citep{bisk2020piqa}: Evaluates physical commonsense reasoning through everyday scenarios, with about 20,000 question-answer pairs.
    
    \item \textbf{RACE-High} \citep{lai2017race}: Contains 69,395 questions from 19,527 passages from English examinations for Chinese high school students.
    
    \item \textbf{RACE-Middle} \citep{lai2017race}: Consists of 28,293 questions from 8,718 passages targeted at middle school students.
    
    \item \textbf{SciQ} \citep{welbl2017sciq}: Contains 13,679 crowdsourced science exam questions covering topics such as physics, chemistry, and biology.
    
    \item \textbf{SIQA} \citep{sap2019socialiqa}: Contains over 38,000 multiple-choice questions about everyday social interactions.
    
    \item \textbf{WinoGrande} \citep{sakaguchi2020winogrande}: Contains 44,000 problems designed to evaluate commonsense reasoning in AI systems.
\end{itemize}

\subsection{Quantifying Conversational Content in Web-Scale Corpora}
\label{sec:quantify_conv}

% \vd{ Updated stats are 
% Total rows: 476276019
% Selected rows: 17484951
% Proportion: 0.0367 (3.67\%)}

To understand the proportion of conversational text in the web data, we conducted an experiment using a classification task on a sample of data from the C4 dataset. The goal was to assess what proportion of naturally occurring web data aligns with chat-based use cases. 

For this, we provided the GPT-4 model with a classification task to label text as either conversational or non-conversational based on the following criteria for classification.

\begin{tcolorbox}[
  title = {Classification Task: Conversational vs.\ Non‑Conversational},
  colback = LightCyan!20,
  colframe = gray!75,
  rounded corners,
  sharp corners = northeast,
  sharp corners = southwest,
  breakable,            % <-- allow page breaks
  enhanced,             % loaded automatically by `breakable`, but explicit is fine
  before skip = 10pt,   % vertical space before the box (optional)
  after  skip = 10pt    % vertical space after the box (optional)
]
\textbf{Your task is to classify text as either conversational (1) or non-conversational (0).}

\textbf{Conversational text (1) shows:}
\begin{itemize}
    \item Back-and-forth exchanges between participants
    \item Questions followed by relevant answers
\end{itemize}

\textbf{Non-conversational text (0) typically:}
\begin{itemize}
    \item Presents information in a one-way format
    \item Lacks interaction between participants
    \item Contains formal or encyclopedic writing
    \item Focuses on describing or explaining without dialogue
\end{itemize}

Rate each text as exactly 0 or 1. Return only the number.

\textbf{Examples:}
\begin{itemize}
    \item \textbf{Text:} "The mitochondria is the powerhouse of the cell. It produces energy through cellular respiration. This process involves multiple steps including glycolysis and the Krebs cycle."\\
    \textbf{Output:} 0
    \item \textbf{Text:} "A: Hey, I'm having trouble with my code. The function keeps returning null. B: Can you share the error message you're getting? A: Here's the stack trace: [error details] B: Ah, I see the issue. You need to initialize the variable first. A: That fixed it, thanks!"\\
    \textbf{Output:} 1
    \item \textbf{Text:} "Q: What's the capital of France? A: The capital of France is Paris. Q: What's its population? A: Paris has a population of about 2.2 million people."\\
    \textbf{Output:} 1
    \item \textbf{Text:} "To install Python, first download the installer from python.org. Run the executable and follow the installation wizard. Make sure to check the 'Add Python to PATH' option during installation."\\
    \textbf{Output:} 0
    \item \textbf{Text:} "JavaScript was created by Brendan Eich in 1995 while he was working at Netscape Communications Corporation. The language was originally designed for client-side web development."\\
    \textbf{Output:} 0
    \item \textbf{Text:} "User1: Did anyone solve the memory leak issue? User2: Yes, it was related to the event listeners User1: How did you fix it? User2: We added a cleanup function in the useEffect hook User1: Got it, I'll try that"\\
    \textbf{Output:} 1
    \item \textbf{Text:} "The Renaissance was a period in European history marking the transition from the Middle Ages to modernity and covering the 15th and 16th centuries."\\
    \textbf{Output:} 0
    \item \textbf{Text:} "- Can we push back the deadline? - I need to check with the team first. When would you need it by? - Would next Friday work? - Yes, that should be fine. I'll update the project timeline."\\
    \textbf{Output:} 1
\end{itemize}

Classify the following text:
\end{tcolorbox}

The experiment was performed on a random sample of 10,000 text examples from the C4 dataset, with a final result indicating that approximately 2.7\% of the sampled text was classified as conversational. This proportion is reflective of the prevalence of conversational data within a large-scale corpus drawn from the web.

\section{Additional Results}

\begin{figure}[!ht]
    \centering
    \includegraphics[width=\textwidth]{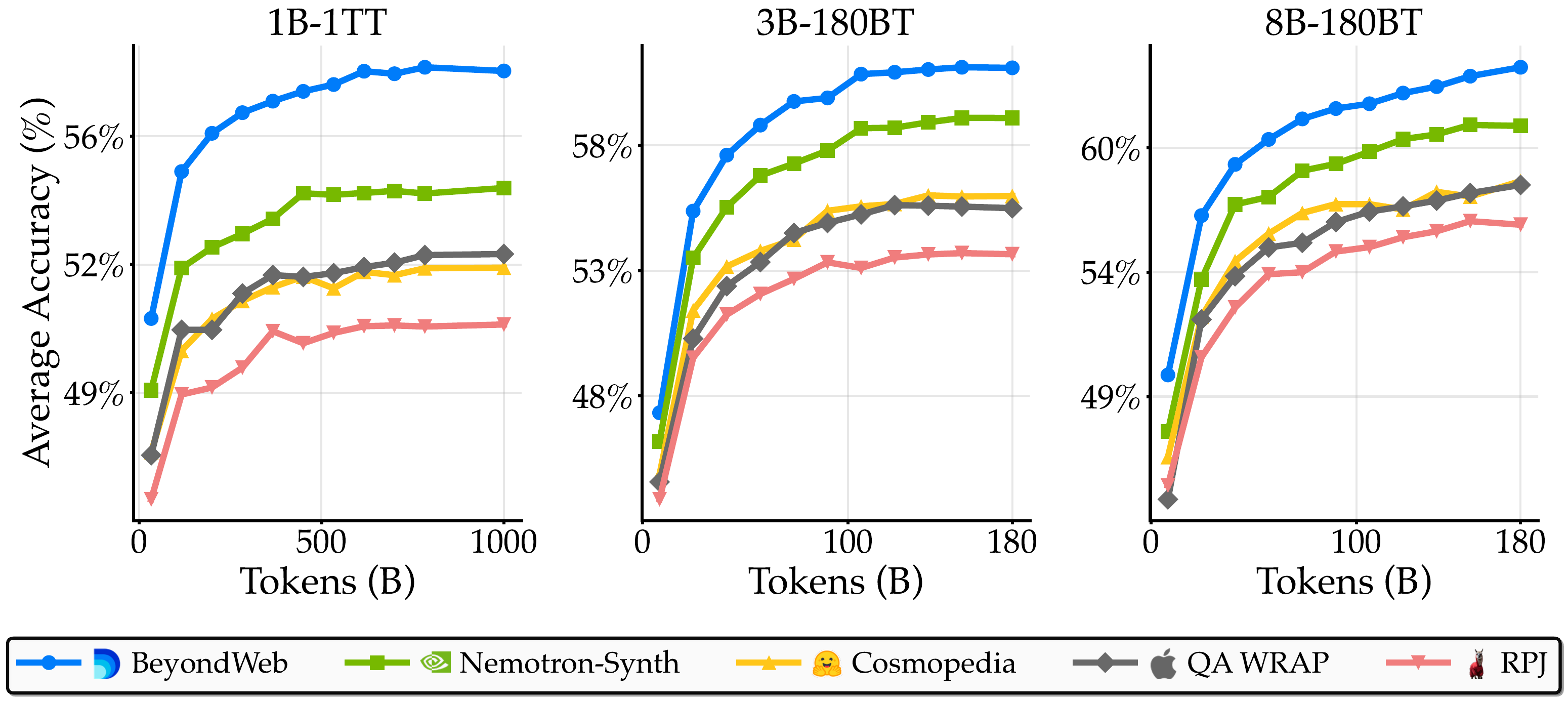}
    \caption{
        Learning curves across model scales demonstrate our consistent performance advantages. The plots show training dynamics for 1B-1TT, 3B-180BT, and 8B-180BT configurations. Average Accuracy (\%) is computed as the mean score across 14 standard benchmarks, averaged over both 0-shot and 5-shot settings. \beyondweb (blue) maintains superior performance throughout training across all scales, achieving 57.4\%, 60.8\%, and 63.7\% accuracy respectively.
    }
    \label{fig:learning_curves}
\end{figure}

\begin{figure}[!ht]
    \centering
    \includegraphics[width=\textwidth]{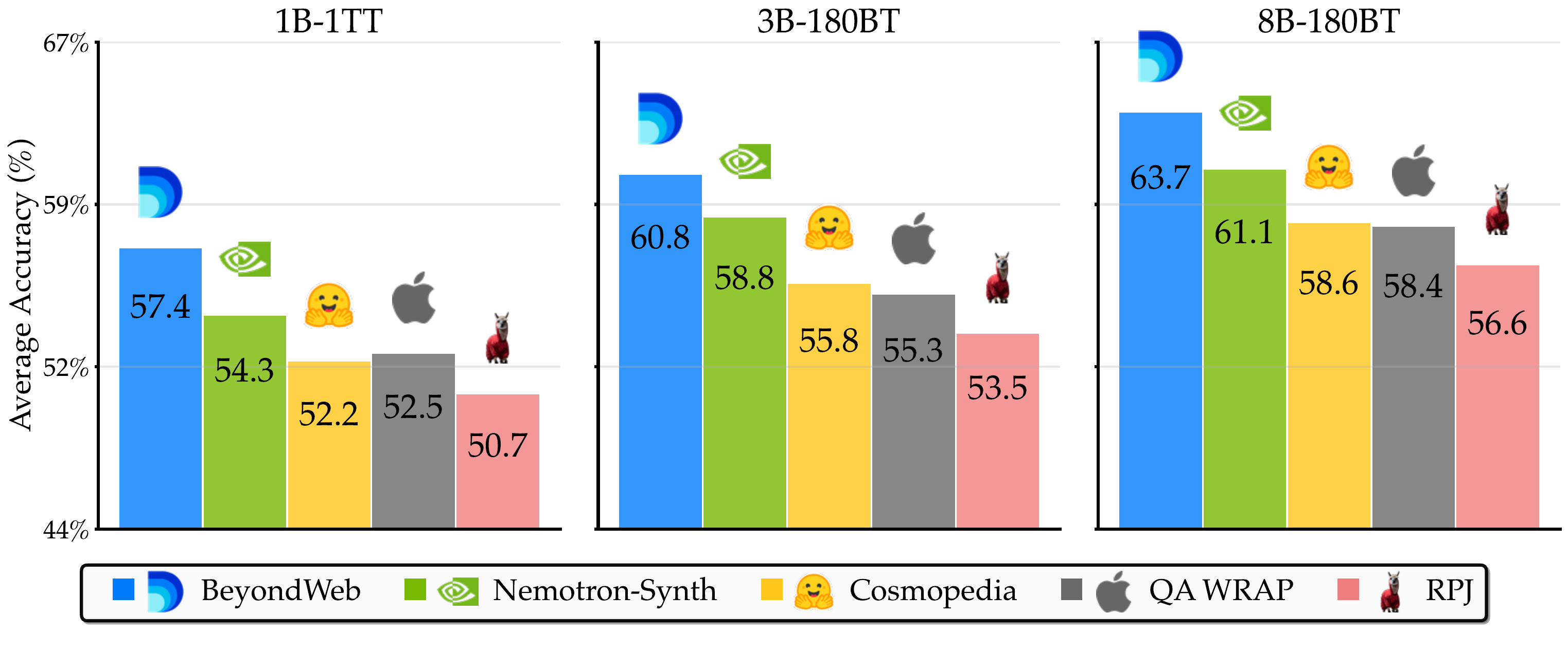}
    \caption{
        Performance comparison showing \beyondweb's significant advantages over state-of-the-art baselines across three model scales. Average Accuracy (\%) is computed as the mean score across 14 standard benchmarks, averaged over both 0-shot and 5-shot settings. We highlight that the 1B \beyondweb model outperforms all but one 3B baseline, and the 3B \beyondweb model outperforms all but one 8B baseline, despite the 3B and 8B models being trained for the same token budget (180B tokens).
    }
    \label{fig:better_performance}
\end{figure}

Figure~\ref{fig:learning_curves} shows that \beyondweb consistently leads throughout training, achieving the highest final accuracy at all scales. As shown in Figure~\ref{fig:better_performance}, the 3B \beyondweb model even surpasses most 8B baselines under the same token budget. Table~\ref{tab:tasks_hero_zeroshot} and Table~\ref{tab:tasks_hero_fiveshot} confirm broad 0-shot and 5-shot improvements across 14 tasks, with \beyondweb outperforming other methods on most of benchmarks at each scale.

\input{tables/hero_0shot}
\input{tables/hero_5shot}

%% file: tables/hero_0shot.tex
% Single table with Scale | Method | Task1 | Task2 | ... | Task15
\begin{table}[!htbp]
\centering
\resizebox{\textwidth}{!}{
\begin{tabular}{ll|ccccccccccccccc}
\toprule
\textbf{Scale} & \textbf{Dataset} & \textbf{ARC(C)} & \textbf{ARC(E)} & \textbf{BoolQ} & \textbf{COPA} & \textbf{CSQA} & \textbf{Hella.} & \textbf{MMLU} & \textbf{OBQA} & \textbf{PIQA} & \textbf{RACE-H} & \textbf{RACE-M} & \textbf{SIQA} & \textbf{SciQ} & \textbf{Wino.} & \textbf{Avg.} \\
\toprule
\multirow{5}{*}{\parbox{1cm}{\centering \textbf{1B} \\ \textbf{(1TT)}}} & RPJ & 27.6 & 50.8 & 60.8 & 73.0 & 38.2 & 51.9 & 31.0 & 34.0 & 71.3 & 35.4 & 43.0 & 42.4 & 75.0 & \underline{54.2} & 49.2 \\
 & QA WRAP & 31.0 & 55.3 & \underline{64.3} & \underline{74.0} & 36.6 & 51.3 & 32.1 & 34.6 & 71.6 & 37.0 & 46.2 & 43.5 & \underline{84.3} & 53.7 & 51.1 \\
 & Cosmopedia & 30.7 & 56.8 & 63.0 & 73.0 & 36.9 & \underline{56.1} & 31.9 & \underline{36.8} & 72.9 & 35.4 & 44.2 & \underline{44.4} & 78.0 & 53.5 & 51.0 \\
 & Nemotron-Synth & \underline{33.4} & \underline{58.8} & 63.2 & 70.0 & \textbf{38.7} & \textbf{56.4} & \underline{33.8} & 36.2 & \underline{73.6} & \underline{39.9} & \textbf{53.1} & 43.2 & 80.7 & 53.8 & \underline{52.5} \\
 & BeyondWeb & \colorbox{lightblue}{\textbf{39.0}} & \colorbox{lightblue}{\textbf{66.4}} & \colorbox{lightblue}{\textbf{67.0}} & \colorbox{lightblue}{\textbf{78.0}} & \underline{38.3} & \colorbox{lightblue}{\textbf{56.4}} & \colorbox{lightblue}{\textbf{34.9}} & \colorbox{lightblue}{\textbf{39.4}} & \colorbox{lightblue}{\textbf{74.6}} & \colorbox{lightblue}{\textbf{41.8}} & \underline{52.9} & \colorbox{lightblue}{\textbf{44.5}} & \colorbox{lightblue}{\textbf{90.3}} & \colorbox{lightblue}{\textbf{56.6}} & \colorbox{lightblue}{\textbf{55.7}} \\
\midrule
\multirow{5}{*}{\parbox{1cm}{\centering \textbf{3B} \\ \textbf{(180BT)}}} & RPJ & 31.2 & 56.4 & 63.0 & 72.0 & 40.0 & 59.0 & 33.5 & 35.8 & 74.4 & 37.7 & 46.7 & 44.2 & 82.1 & 55.7 & 52.3 \\
 & QA WRAP & 34.6 & 59.9 & \underline{65.6} & 70.0 & 39.6 & 58.7 & 33.9 & 36.8 & 72.6 & 39.8 & 50.6 & 44.2 & 86.7 & 55.1 & 53.4 \\
 & Cosmopedia & 34.3 & 60.0 & 62.6 & \underline{73.0} & 40.1 & 63.0 & 34.9 & 38.6 & 75.8 & 38.6 & 48.5 & \underline{44.7} & 82.6 & 56.8 & 53.8 \\
 & Nemotron-Synth & \underline{38.0} & \underline{68.5} & 65.0 & \textbf{80.0} & \underline{41.3} & \textbf{64.2} & \textbf{37.2} & \underline{39.6} & \underline{76.6} & \underline{43.2} & \underline{56.1} & 43.8 & \underline{88.7} & \underline{58.2} & \underline{57.2} \\
 & BeyondWeb & \colorbox{lightblue}{\textbf{44.0}} & \colorbox{lightblue}{\textbf{71.5}} & \colorbox{lightblue}{\textbf{67.4}} & \colorbox{lightblue}{\textbf{80.0}} & \colorbox{lightblue}{\textbf{42.0}} & \underline{63.7} & \underline{36.9} & \colorbox{lightblue}{\textbf{41.6}} & \colorbox{lightblue}{\textbf{76.9}} & \colorbox{lightblue}{\textbf{44.7}} & \colorbox{lightblue}{\textbf{58.1}} & \colorbox{lightblue}{\textbf{45.9}} & \colorbox{lightblue}{\textbf{91.8}} & \colorbox{lightblue}{\textbf{58.8}} & \colorbox{lightblue}{\textbf{58.8}} \\
\midrule
\multirow{5}{*}{\parbox{1cm}{\centering \textbf{8B} \\ \textbf{(180BT)}}} & RPJ & 33.3 & 61.9 & 64.2 & 74.0 & 42.1 & 65.9 & 35.3 & 38.6 & 76.2 & 39.5 & 47.9 & 44.7 & 83.9 & 59.0 & 54.7 \\
 & QA WRAP & 37.9 & 64.2 & \underline{68.8} & 77.0 & 41.8 & 64.6 & 36.8 & 36.4 & 74.7 & 41.1 & 52.9 & 44.0 & \underline{90.0} & 58.5 & 56.3 \\
 & Cosmopedia & 38.1 & 65.3 & 64.6 & \underline{80.0} & 41.9 & 68.6 & 37.0 & 40.2 & \underline{77.4} & 41.3 & 52.4 & \underline{45.0} & 84.3 & 58.6 & 56.8 \\
 & Nemotron-Synth & \underline{43.3} & \underline{72.3} & 65.4 & \textbf{82.0} & \underline{43.8} & \underline{69.1} & \underline{38.9} & \underline{41.4} & 77.3 & \underline{44.4} & \underline{57.5} & 44.8 & 89.1 & \underline{59.5} & \underline{59.2} \\
 & BeyondWeb & \colorbox{lightblue}{\textbf{48.9}} & \colorbox{lightblue}{\textbf{76.5}} & \colorbox{lightblue}{\textbf{73.0}} & \underline{80.0} & \colorbox{lightblue}{\textbf{45.1}} & \colorbox{lightblue}{\textbf{69.2}} & \colorbox{lightblue}{\textbf{39.6}} & \colorbox{lightblue}{\textbf{43.6}} & \colorbox{lightblue}{\textbf{78.8}} & \colorbox{lightblue}{\textbf{47.1}} & \colorbox{lightblue}{\textbf{60.0}} & \colorbox{lightblue}{\textbf{46.2}} & \colorbox{lightblue}{\textbf{94.4}} & \colorbox{lightblue}{\textbf{61.9}} & \colorbox{lightblue}{\textbf{61.7}} \\
\bottomrule
\end{tabular}
}
\caption{Performance across all tasks (0-shot) for different model scales and data curations. The best value at each scale is highlighted in \textbf{bold}, the second-best is \underline{underlined}, and results where \beyondweb outperforms all other methods are additionally shaded in \colorbox{lightblue}{blue}.
}
\label{tab:tasks_hero_zeroshot}
\end{table}

%% file: tables/hero_5shot.tex
% Single table with Scale | Method | Task1 | Task2 | ... | Task15
\begin{table}[!htbp]
\centering
\resizebox{\textwidth}{!}{
\begin{tabular}{ll|ccccccccccccccc}
\toprule
\textbf{Scale} & \textbf{Dataset} & \textbf{ARC(C)} & \textbf{ARC(E)} & \textbf{BoolQ} & \textbf{COPA} & \textbf{CSQA} & \textbf{Hella.} & \textbf{MMLU} & \textbf{OBQA} & \textbf{PIQA} & \textbf{RACE-H} & \textbf{RACE-M} & \textbf{SIQA} & \textbf{SciQ} & \textbf{Wino.} & \textbf{Avg.} \\
\toprule
\multirow{5}{*}{\parbox{1cm}{\centering \textbf{1B} \\ \textbf{(1TT)}}} & RPJ & 31.1 & 60.9 & 59.2 & 71.0 & 49.4 & 51.7 & 32.1 & 36.4 & 71.2 & 35.5 & 43.2 & 44.1 & 90.5 & 53.5 & 52.1 \\
 & QA WRAP & 30.8 & 64.0 & \underline{67.1} & \textbf{75.0} & 51.8 & 50.9 & 33.3 & 35.8 & 71.9 & 37.0 & 47.8 & 45.1 & 92.2 & 52.8 & 54.0 \\
 & Cosmopedia & 37.4 & 68.9 & 56.3 & 68.0 & 50.9 & 55.1 & 33.7 & \underline{37.2} & 73.7 & 33.6 & 42.7 & \underline{46.3} & 90.8 & 52.6 & 53.4 \\
 & Nemotron-Synth & \underline{38.4} & \underline{69.6} & 62.4 & 73.0 & \underline{55.5} & \underline{55.9} & \underline{35.3} & 36.8 & \underline{74.6} & \underline{38.7} & \underline{52.0} & 45.3 & \underline{93.1} & \underline{54.4} & \underline{56.1} \\
 & BeyondWeb & \colorbox{lightblue}{\textbf{43.9}} & \colorbox{lightblue}{\textbf{75.1}} & \colorbox{lightblue}{\textbf{69.6}} & \colorbox{lightblue}{\textbf{75.0}} & \colorbox{lightblue}{\textbf{60.2}} & \colorbox{lightblue}{\textbf{55.9}} & \colorbox{lightblue}{\textbf{36.1}} & \colorbox{lightblue}{\textbf{38.8}} & \colorbox{lightblue}{\textbf{75.4}} & \colorbox{lightblue}{\textbf{41.6}} & \colorbox{lightblue}{\textbf{54.5}} & \colorbox{lightblue}{\textbf{50.4}} & \colorbox{lightblue}{\textbf{95.6}} & \colorbox{lightblue}{\textbf{55.3}} & \colorbox{lightblue}{\textbf{59.1}} \\
\midrule
\multirow{5}{*}{\parbox{1cm}{\centering \textbf{3B} \\ \textbf{(180BT)}}} & RPJ & 34.8 & 66.1 & 60.0 & 68.0 & 55.9 & 58.1 & 34.2 & 36.4 & 74.8 & 37.8 & 46.8 & 45.2 & 92.7 & 54.5 & 54.7 \\
 & QA WRAP & 37.4 & 69.0 & \underline{68.9} & 76.0 & 53.5 & 58.4 & 36.1 & 36.2 & 73.8 & 40.4 & 52.8 & 48.2 & 93.6 & 55.4 & 57.1 \\
 & Cosmopedia & 40.8 & 72.2 & 63.6 & 76.0 & 57.7 & 61.6 & 36.2 & 39.6 & 76.2 & 38.2 & 50.1 & 47.8 & 93.4 & 54.1 & 57.7 \\
 & Nemotron-Synth & \underline{44.4} & \underline{74.2} & 67.6 & \underline{81.0} & \underline{59.1} & \textbf{63.7} & \underline{38.2} & \underline{43.2} & \underline{77.1} & \underline{42.1} & \underline{53.9} & \underline{49.2} & \underline{95.8} & \textbf{57.1} & \underline{60.5} \\
 & BeyondWeb & \colorbox{lightblue}{\textbf{49.3}} & \colorbox{lightblue}{\textbf{79.2}} & \colorbox{lightblue}{\textbf{74.1}} & \colorbox{lightblue}{\textbf{84.0}} & \colorbox{lightblue}{\textbf{61.3}} & \underline{63.3} & \colorbox{lightblue}{\textbf{38.6}} & \colorbox{lightblue}{\textbf{44.4}} & \colorbox{lightblue}{\textbf{77.2}} & \colorbox{lightblue}{\textbf{44.5}} & \colorbox{lightblue}{\textbf{57.6}} & \colorbox{lightblue}{\textbf{53.0}} & \colorbox{lightblue}{\textbf{96.1}} & \underline{56.2} & \colorbox{lightblue}{\textbf{62.8}} \\
\midrule
\multirow{5}{*}{\parbox{1cm}{\centering \textbf{8B} \\ \textbf{(180BT)}}} & RPJ & 36.8 & 69.9 & 67.9 & 77.0 & 60.6 & 65.5 & 36.7 & 40.4 & 76.3 & 39.7 & 47.6 & 49.3 & 94.6 & 56.5 & 58.5 \\
 & QA WRAP & 41.9 & 72.7 & \underline{72.3} & 81.0 & 59.0 & 64.0 & 38.9 & 40.2 & 76.4 & 41.9 & 56.1 & 50.0 & 95.4 & 56.8 & 60.5 \\
 & Cosmopedia & 44.9 & 76.1 & 64.0 & 79.0 & 59.3 & \underline{67.7} & 38.5 & \underline{43.0} & 77.3 & 41.2 & 51.8 & 49.6 & 94.8 & \underline{58.2} & 60.4 \\
 & Nemotron-Synth & \underline{48.9} & \underline{79.3} & 70.4 & \underline{85.0} & \underline{61.0} & \textbf{69.0} & \underline{41.1} & 42.2 & \underline{78.0} & \underline{43.8} & \underline{57.5} & \underline{50.1} & \underline{96.0} & \underline{58.2} & \underline{62.9} \\
 & BeyondWeb & \colorbox{lightblue}{\textbf{54.7}} & \colorbox{lightblue}{\textbf{82.1}} & \colorbox{lightblue}{\textbf{78.1}} & \colorbox{lightblue}{\textbf{86.0}} & \colorbox{lightblue}{\textbf{63.6}} & \colorbox{lightblue}{\textbf{69.0}} & \colorbox{lightblue}{\textbf{41.5}} & \colorbox{lightblue}{\textbf{47.6}} & \colorbox{lightblue}{\textbf{78.3}} & \colorbox{lightblue}{\textbf{46.9}} & \colorbox{lightblue}{\textbf{60.9}} & \colorbox{lightblue}{\textbf{54.4}} & \colorbox{lightblue}{\textbf{96.8}} & \colorbox{lightblue}{\textbf{59.0}} & \colorbox{lightblue}{\textbf{65.6}} \\
\bottomrule
\end{tabular}
}
\caption{Performance across all tasks (5-shot) for different model scales and data curations. The best value at each scale is highlighted in \textbf{bold}, the second-best is \underline{underlined}, and results where \beyondweb outperforms all other methods are additionally shaded in \colorbox{lightblue}{blue}.
}
\label{tab:tasks_hero_fiveshot}
\end{table}

% \caption{Performance across all tasks (5-shot) for different model scales and data curations. The best value at each scale is highlighted in \textbf{bold}, the second-best is \underline{underlined}, and results where \beyondweb outperforms all other methods are additionally shaded in \colorbox{lightblue}{blue}.
% }